\documentclass[11pt, a4paper, logo, copyright, nonumbering]{deepseek}
\usepackage[authoryear, compress, round]{natbib} 
\usepackage{dblfloatfix}
\usepackage{ulem}
\usepackage{caption}
\usepackage{dramatist}
\usepackage{xspace}
\usepackage{pifont}
\usepackage{multirow}
\usepackage{tcolorbox}
\usepackage{xltabular}
\usepackage{diagbox}
\usepackage{longtable}
\usepackage{hyperref}
\usepackage{algorithm}
\usepackage{algpseudocode}
\usepackage{amsfonts}
\usepackage{amsmath}
\usepackage{amssymb}
\usepackage{lineno}
\usepackage{multirow}
\usepackage{adjustbox}
\usepackage{listings}
\usepackage[bottom]{footmisc}

\usepackage{CJKutf8}
\usepackage{setspace}

\usepackage{dsfont}
\usepackage{array}
\usepackage{tabularx}
\usepackage{subfigure}
\usepackage{xcolor}
\usepackage{listings}

\usepackage{lipsum}
\usepackage{multicol}

\makeatletter
\def\@BTrule[#1]{
  \ifx\longtable\undefined
    \let\@BTswitch\@BTnormal
  \else\ifx\hline\LT@hline
    \nobreak
    \let\@BTswitch\@BLTrule
  \else
     \let\@BTswitch\@BTnormal
  \fi\fi
  \global\@thisrulewidth=#1\relax
  \ifnum\@thisruleclass=\tw@\vskip\@aboverulesep\else
  \ifnum\@lastruleclass=\z@\vskip\@aboverulesep\else
  \ifnum\@lastruleclass=\@ne\vskip\doublerulesep\fi\fi\fi
  \@BTswitch}
\makeatother

\addto\extrasenglish{

}

\reportnumber{001}

\renewcommand{\today}{}
\newcommand{\mhc}{\textit{m}HC}

\title{
\vspace*{-0.6cm}
\centering
DeepSeek-V4.1-Flash: \\ Pushing the Limits of KV Cache Compression
}

\author[*]{
\vspace{-0.8cm}
DeepSeek-AI
\\
\small
\vspace{-0.3cm}
\texttt{research@deepseek.com}
\vspace{-0.9cm}
}

\renewcommand{\phi}{\varphi}

\renewcommand{\leq}{\leqslant}
\renewcommand{\geq}{\geqslant}

\renewcommand{\epsilon}{\varepsilon}
\renewcommand{\imath}{\mathrm{i}}

\newlength{\restsubwidth}
\newlength{\restsubheight}
\newlength{\restsubmoreheight}
\newcommand{\rest}[2]{%
        \settowidth{\restsubwidth}{\ensuremath{#2}}
        \settoheight{\restsubheight}{\ensuremath{{}_{#2}}}
        \ensuremath{{#1\hskip 0.5pt}_{\vrule\kern2pt\parbox[b][%
        4pt][b]{\the\restsubwidth}{%
                        \ensuremath{{}_{#2}}}}}
        }

\begin{abstract}
\vspace{-0.5cm}

The widespread adoption of long-horizon agents has made model workloads increasingly input-heavy. Although prior work has substantially reduced the cost of long-context computation, prefill remains computationally expensive, and large KV caches continue to strain HBM and SSD capacity and data-transfer bandwidth. Together, these compute, storage, and bandwidth demands constitute the primary bottleneck to further lowering deployment costs. To address this challenge, we introduce DeepSeek-V4.1-Flash, a multimodal Mixture-of-Experts (MoE) model with 552B backbone parameters and support for contexts of up to one million tokens. With its Causal Encoder-Decoder (CED) architecture, the model activates 16B parameters per token during decode but only 8B parameters during prefill, substantially improving cost efficiency for agentic workloads.
To push the limits of KV cache compression, DeepSeek-V4.1-Flash combines cross-layer KV cache reuse in Compressed Sparse Attention 2 (CSA2) with FP4 KV caching.
These designs reduce its global KV cache footprint (always in HBM) to 890 bytes per token, roughly 1/4 of the corresponding footprint of DeepSeek-V4-Flash.
Further, through a dedicated deployment optimization known as SWA Bounded Replay, DeepSeek-V4.1-Flash reduces its persistent KV cache footprint (always on SSD or in host memory) to roughly 1/8 of that of DeepSeek-V4-Flash.
Despite its much smaller KV cache footprint, the model delivers substantially better performance than the baseline.
In addition, we streamline the DeepSeek-V4 architecture and introduce several efficient architectural extensions.
We pretrain DeepSeek-V4.1-Flash on a multimodal corpus comprising 45T tokens and conduct comprehensive post-training, yielding strong performance across diverse text-based and multimodal agentic scenarios. Model checkpoints are available at \href{https://huggingface.co/deepseek-ai/DeepSeek-V4.1-Flash}{https://huggingface.co/deepseek-ai/DeepSeek-V4.1-Flash}.

\end{abstract}

\begin{document}
\begin{CJK*}{UTF8}{gbsn}

\maketitle

\begin{figure}[h]
    \centering
    \begingroup
    \setlength{\unitlength}{\textwidth}
    \begin{picture}(1,0.365)
        \put(0,0.025){
            \includegraphics[width=0.49\textwidth]
                {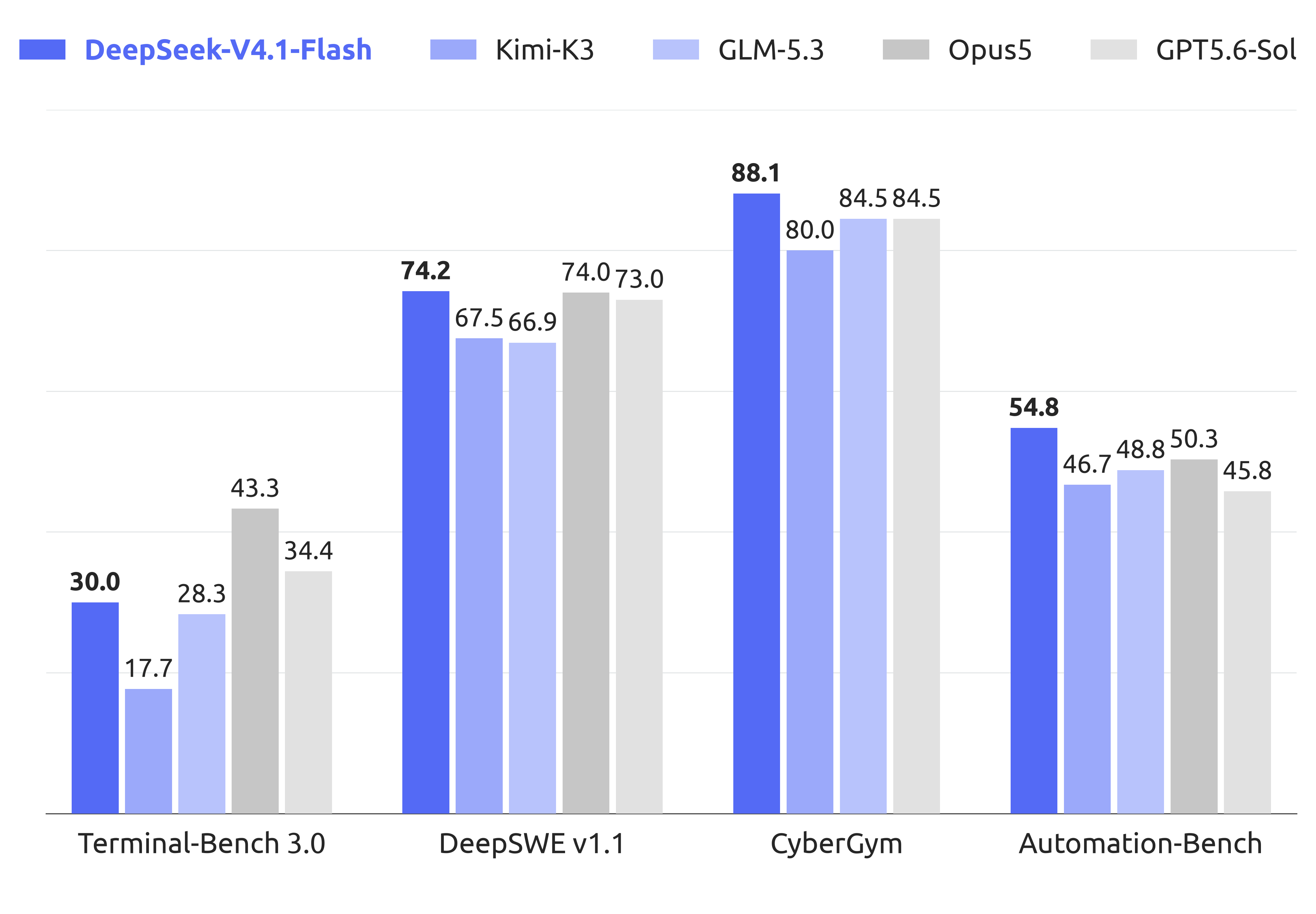}
        }
        \put(0.51,0.025){
            \includegraphics[width=0.49\textwidth]
                {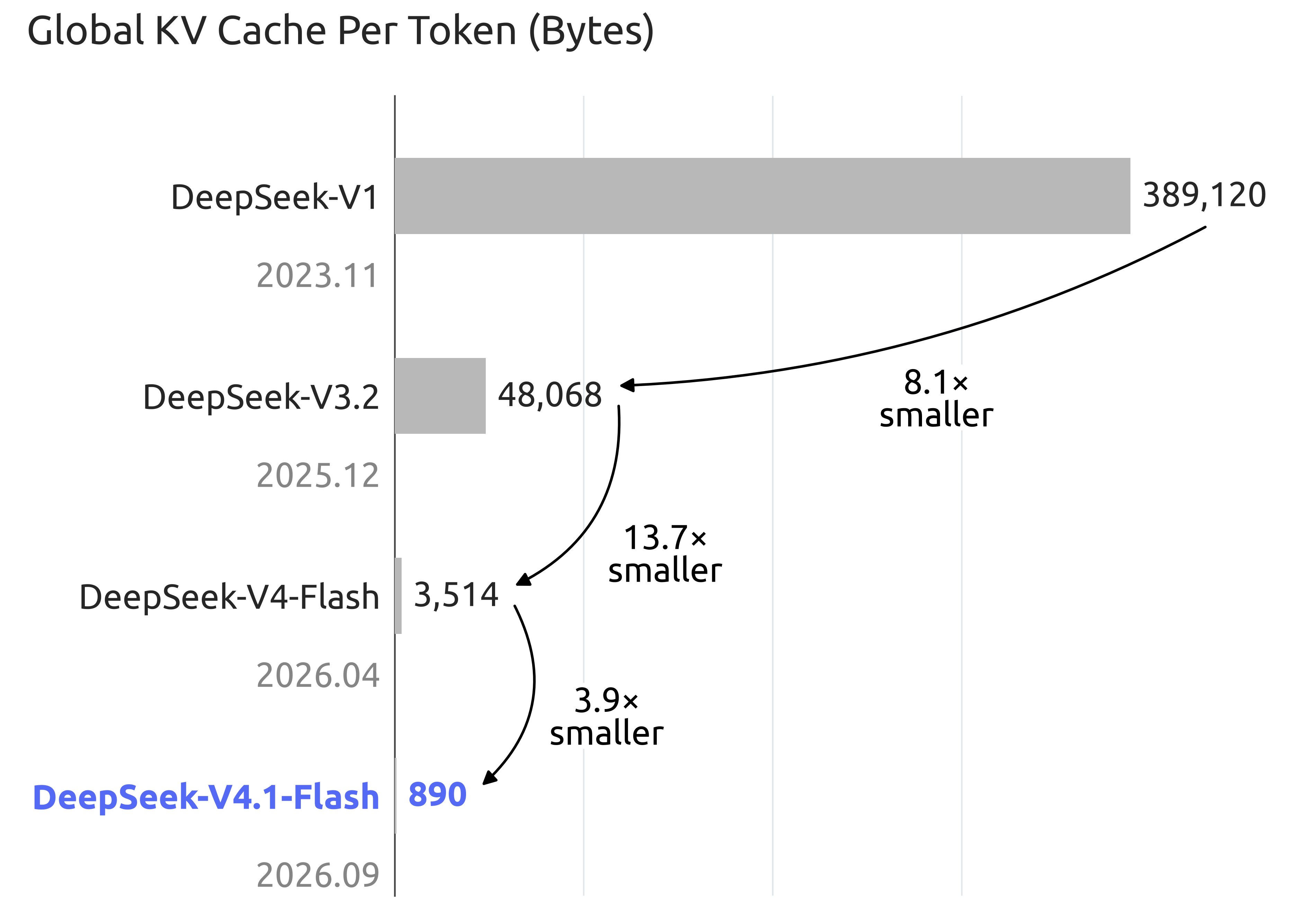}
        }
        \put(0.245,0.003){
            \makebox(0,0)[b]{\scriptsize\textbf{(a)}}
        }
        \put(0.755,0.003){
            \makebox(0,0)[b]{\scriptsize\textbf{(b)}}
        }
    \end{picture}
    \endgroup
    \caption{
        \textbf{(a)} Performance of DeepSeek-V4.1-Flash and its counterparts on agentic benchmarks.
        \textbf{(b)} Global KV cache size per token (in bytes) across generations of DeepSeek models, highlighting DeepSeek’s sustained efforts to reduce context memory requirements. DeepSeek-V4.1-Flash achieves approximately 4-fold and 437-fold reductions in per-token global KV cache size relative to DeepSeek-V4-Flash and DeepSeek-V1, respectively.
    }
    \label{fig:teaser}
\end{figure}

\newpage

\begin{spacing}{0.9}
\tableofcontents
\end{spacing}

\newpage

\section{Introduction}

Applications of long-horizon agents have expanded rapidly in recent years, making ultra-long-context processing an increasingly important model workload. Supporting such workloads requires not only efficient long-sequence processing, but also the persistent storage, reuse, and transfer of large KV caches. KV cache management has therefore become a foundational capability for model deployment, while introducing substantial challenges across computation, storage, and communication. Prior advances in sparse attention~\citep{dsv32,dsv4} have significantly reduced the computational cost of long-sequence processing, making persistent storage and data movement increasingly prominent bottlenecks.

Specifically, DeepSeek-V4~\citep{dsv4} combines a global attention branch spanning the full context with local Sliding-Window Attention (SWA). The global branch maintains global KV, comprising main KV and indexer K, while SWA maintains local KV states. For a fixed window size, SWA KV storage is bounded independently of sequence length. For sufficiently long sequences, global KV therefore dominates the \textbf{runtime KV} footprint, which is constrained by HBM capacity. In addition, certain KV are persisted for prefix reuse, referred to as \textbf{persistent KV} caches, which are constrained by SSD and host memory capacity. I/O and interconnect bandwidth also limit cache migration and loading. Together, these constraints limit serving throughput, increase deployment costs, and ultimately hinder the deployment and adoption of agents over longer task horizons and across broader application scenarios.

Further reducing the KV cache footprint is therefore critical to alleviating storage and communication bottlenecks and lowering the cost of long-context serving. To this end, we develop DeepSeek-V4.1-Flash, a multimodal Mixture-of-Experts (MoE) model designed for more aggressive KV cache compression. DeepSeek-V4.1-Flash has 552B backbone parameters, natively supports multimodal inputs, and accommodates contexts of up to one million tokens. We adopt a Causal Encoder-Decoder (CED) architecture, in which decoder global KV is projected from the final encoder hidden states. This design enables the model to activate 8B parameters per token during prefill and 16B during decode, which is particularly cost-effective for input-heavy agentic scenarios. Despite being considerably larger than DeepSeek-V4-Flash, DeepSeek-V4.1-Flash requires only approximately 1/4 as much runtime KV cache storage and 1/8 as much persistent KV cache storage at the same sequence length. Moreover, DeepSeek-V4.1-Flash delivers better overall performance than DeepSeek-V4-Flash.

This level of KV cache compression is achieved through joint optimizations in model architecture, cache precision, and deployment strategy. Conceptually, DeepSeek-V4 can be viewed as an SWA-based local-processing backbone augmented with compressed global context. This perspective motivates us to focus on simplifying the global branch while largely preserving the local attention design. At the architectural level, we design Compressed Sparse Attention 2 (CSA2), which applies cross-layer reuse to global KV (including main KV and indexer K) and Top-K indices to substantially reduce KV cache storage. 
CSA2 has three statically assigned modes: Full, Reindex, and Reuse. Full Mode generates global KV and performs indexing. Reindex Mode reuses the global KV from a preceding layer, and uses its own indexer Q to rescore the shared indexer K and select fresh Top-K indices. Reuse Mode reuses both global KV and the Top-K indices in a preceding layer, and directly performs sparse attention. In all three modes, each layer retains its own global Q and SWA KV. Sharing global KV and indexer K reduces duplicated cache storage. In addition, different from DeepSeek-V4 that employs the Compressed Sparse Attention (CSA)--Heavily Compressed Attention (HCA) hybrid architecture, DeepSeek-V4.1-Flash uses pure CSA2.
At the cache-precision level, we use FP4 global KV caches during training with only marginal performance degradation. Together, CSA2 and FP4 KV caching reduce global KV cache storage to approximately 1/4 of that of DeepSeek-V4-Flash, as shown in Figure~\ref{fig:teaser}(b). At the deployment level, DeepSeek-V4.1-Flash, like DeepSeek-V4, uses Sliding-Window Attention (SWA) in every layer. In DeepSeek-V4, we use a hybrid strategy to balance the storage cost of persisting SWA KV caches against the computation required for exact reconstruction. Exact reconstruction requires replaying the most recent $L \times n_{\mathrm{win}}$ tokens, where $L$ is the number of layers and $n_{\mathrm{win}}$ is the SWA window size. In DeepSeek-V4.1-Flash, we introduce SWA Bounded Replay, which approximately reconstructs the required SWA KV states by replaying only the most recent $n_{\mathrm{win}}$ tokens. Our experiments show that this incurs only negligible performance degradation. This finding establishes a new storage--computation trade-off, allowing us to avoid persisting SWA KV cache to SSD while incurring a small amount of prefill recomputation. With SWA Bounded Replay, the persistent KV cache footprint is further reduced to approximately 1/8 of that of DeepSeek-V4-Flash. Together, these optimizations greatly ease pressure on HBM and SSD capacity, reduce deployment costs, and pave the way for deployment at a larger scale.

\begin{figure}[t]
\centering
\includegraphics[width=0.6\textwidth]{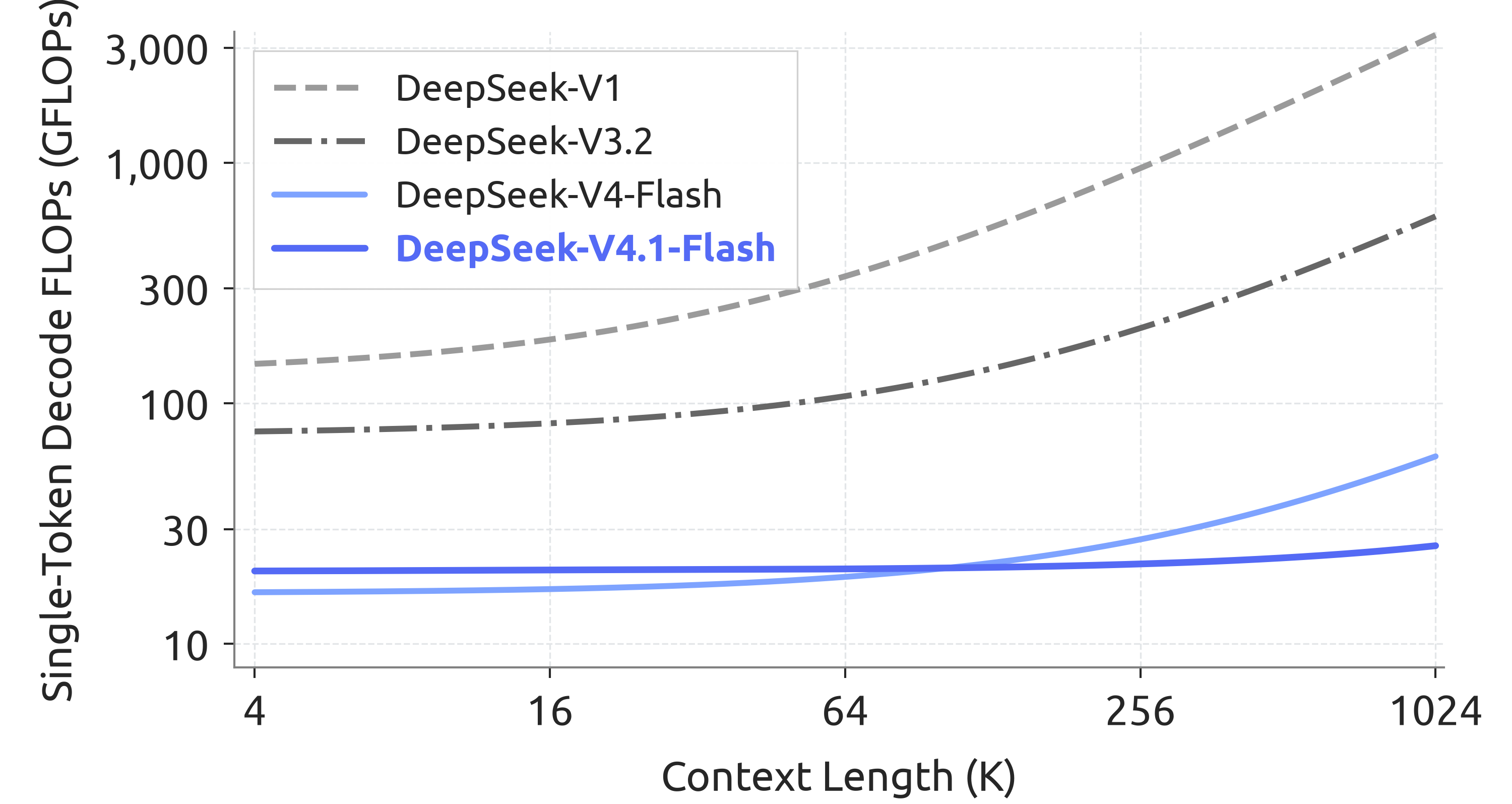}
\caption{Single-token Decode FLOPs versus context length across generations of DeepSeek models. We account for compute precision by weighting BF16, FP8, and FP4 operations by $1$, $0.5$, and $0.25$, respectively. DeepSeek-V4.1-Flash maintains nearly constant Decode FLOPs as context length increases, substantially reducing the computational cost of long-context scenarios.
}
\label{fig:decode_flops_curves}
\end{figure}

Complementing CED and CSA2, we further streamline the original DeepSeek-V4 architecture. Additionally, we upgrade the original \mhc{}~\citep{mhc} design to Single-Pass \mhc{}, with an accompanying Mega-\mhc{} deployment kernel that halves activation memory traffic relative to the original four-kernel implementation. Furthermore, we integrate the Engram~\citep{engram} conditional memory module to strengthen model capabilities. We also introduce the DSpark~\citep{cheng2026dspark} speculative decoding architecture to improve decoding efficiency through semi-autoregressive draft generation and confidence-scheduled verification.
With all the architectural improvements combined, the single-token Decode FLOPs of DeepSeek-V4.1-Flash remain nearly constant across context lengths. Figure~\ref{fig:decode_flops_curves} shows that extending the context length 256-fold, from 4K to 1M, increases its Decode FLOPs by only 1/4, significantly less than the growth observed for DeepSeek-V4-Flash.

To fully realize the KV cache compression benefits of these architectural designs and further improve training and inference efficiency, we systematically co-optimize the training infrastructure and inference system for DeepSeek-V4.1-Flash, ensuring efficient and scalable large-scale multimodal training and long-context deployment. Training infrastructure supports disaggregated vision-encoder execution, balanced image sharding for long sequences, and cross-stage shared-state management for attention reuse. The inference system implements Encoder and Decoder SWA Bounded Replay paths.
Further optimizations include communication--computation overlap, sharded Engram embedding tables, and inference kernel fusion. In particular, each CSA2 Reuse Mode layer executes with only 15 kernels during prefill and 11 during decode. We also separate long-lived global KV storage from short-lived encoder SWA KV in host memory, using bounded replay to approximately reconstruct missing encoder SWA states.

During pre-training, we train DeepSeek-V4.1-Flash on a large-scale multimodal corpus comprising 45T tokens. Sparse attention is trained from scratch at a sequence length of 64K, without any dense attention warmup stages. After pre-training, the model possesses native multimodal capabilities and supports contexts of up to one million tokens.
In our evaluations, DeepSeek-V4.1-Flash-Base achieves world knowledge, reasoning and coding abilities comparable to DeepSeek-V4-Pro-Base, and delivers 5\%--10\% improvements on held-out evaluations, using only 1/3 total parameters and 1/4 activated parameters.
Together, these results highlight its strong parameter efficiency and reflect improvements in training data quality for real-world deployment.

Building on this base model, we conduct post-training to elicit its reasoning and agentic capabilities. In contrast to the architectural innovations described above, our post-training introduces no algorithmic innovation: the recipe follows the standard paradigm of supervised fine-tuning (SFT) followed by reinforcement learning (RL) and on-policy distillation (OPD), without any modification beyond well-established practice used in DeepSeek-V4 development \citep{dsv4}. All substantive changes lie instead in the data pipeline. We develop large-scale automated pipelines for data synthesis and environment construction, and progressively scale the data, tasks, and rollouts employed during RL, thereby extending the model's capabilities across textual, multimodal, and agentic domains.
Figure~\ref{fig:teaser}(a) summarizes DeepSeek-V4.1-Flash's performance on core agentic benchmarks. Our evaluation shows that, despite its compact size, DeepSeek-V4.1-Flash exhibits a distinctive capability profile:
\begin{itemize}
    \item \textbf{Reasoning.} The model delivers strong reasoning ability, sustaining high accuracy on reasoning-intensive benchmarks such as mathematics and competitive programming, showing comparable performance with top open-source models, such as Kimi-K3\citep{kimi_k3} and DeepSeek-V4-Pro.
    \item \textbf{Agent.} 
    DeepSeek-V4.1-Flash achieves performance on par with closed-source frontier models across standard agentic benchmarks like Terminal-Bench 2.1~\citep{merrill2026terminal}, DeepSWE v1.1~\citep{deepswe}, and AutomationBench~\citep{shepard2026automationbench}. It has proven fully capable of handling everyday coding tasks and white-collar workflows. 
    However, a gap with giant models remains on science-oriented agentic tasks, such as Terminal-Bench 4.0~\citep{Marten_Terminal-Bench_2026}, that require expert-level domain knowledge.
    \item \textbf{Multimodal.} Within the multimodal domain, the model surpasses top-tier open-source competitors like Kimi-K3 specifically on benchmarks evaluating visual reasoning and the interpretation of professional charts. Beyond formal metrics, it also exhibits practical utility in real-world visual agentic workflows, such as frontend development and office automation, where it can utilize rendered screen captures for visual inspection and self-correction. Nevertheless, we acknowledge that a distinct overall performance gap remains when compared to giant closed-source systems.
\end{itemize}
These results indicate that DeepSeek-V4.1-Flash can already match closed-source frontier models on the vast majority of benchmarks, and is capable of completing over 95\% of real-world tasks. Meanwhile, its small activation footprint yields low inference latency and serving cost. We therefore believe that DeepSeek-V4.1-Flash offers a favorable trade-off between capability and efficiency, and can serve as a fast, affordable assistant supporting the daily work of a broad population of users. In summary, DeepSeek-V4.1-Flash simultaneously improves model intelligence and inference efficiency while reducing deployment costs. It substantially lowers the cost barrier to deploying long-horizon agents at scale and creates new opportunities for their adoption across a broader range of scenarios. DeepSeek-V4.1-Flash also serves as a new starting point for our continued scaling efforts. Building on this foundation, we will pursue the joint scaling of model architecture, pre-training, and post-training to further explore the frontier of model intelligence.

\begin{figure}[t]
\centering
\includegraphics[width=\textwidth]{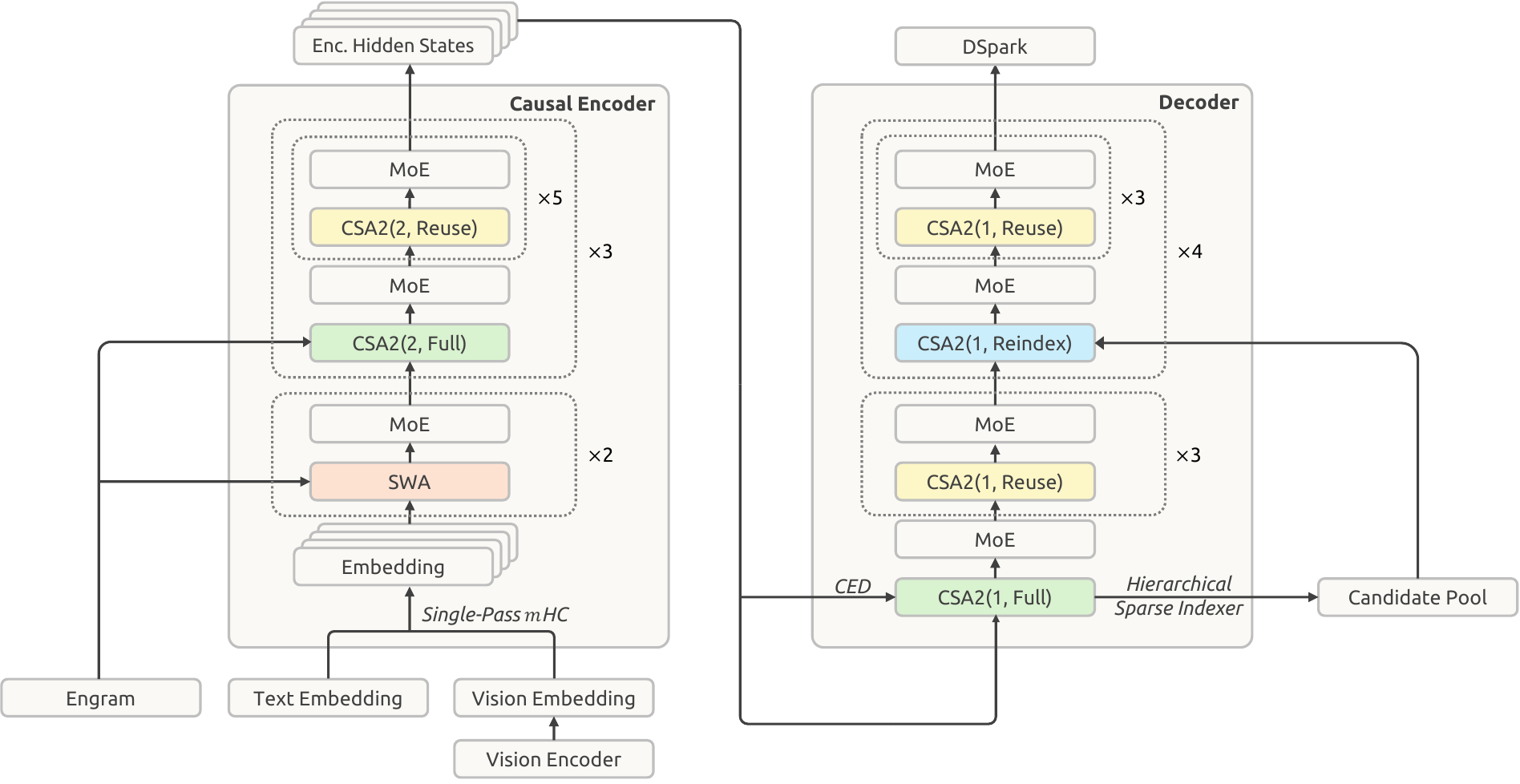}
\caption{
\textbf{Overall architecture of DeepSeek-V4.1-Flash.}
The 40-layer network is divided into a causal encoder and a decoder, each with 20 layers.
All feed-forward layers use standard DeepSeekMoE.
The first two encoder layers use sliding window attention (SWA); the rest use Compressed Sparse Attention 2 (CSA2), with CSA2(ratio, mode) specifying the compression ratio and mode.
The model also uses Single-Pass \mhc{}, Engram, DSpark, and a Hierarchical Sparse Indexer.
}
\label{fig:arch_full}
\end{figure}

\section{Architecture}
\label{sec:arch}

\subsection{Overview}

DeepSeek-V4.1-Flash is a multimodal mixture-of-experts (MoE) Transformer that takes images and text as input and generates text autoregressively. Its language backbone comprises 40 causal Transformer layers, organized into a 20-layer causal encoder followed by a 20-layer decoder. Each layer incorporates both global attention and sliding window attention (SWA), except for the first two layers, which use SWA only. A vision encoder and an MLP projector convert images into visual embeddings that are processed jointly with text embeddings, with multimodal data incorporated from the start of language-model pre-training. Overall, DeepSeek-V4.1-Flash has 552B backbone parameters and 196B Engram parameters, activating 8B parameters per token during prefill and 16B during decode. Figure~\ref{fig:arch_full} illustrates the overall architecture of DeepSeek-V4.1-Flash.

The Causal Encoder–Decoder (CED) architecture and Compressed Sparse Attention 2 (CSA2) address complementary costs of long-context inference. CED constructs the decoder's global key-value (KV) cache from encoder outputs, allowing most prompt tokens to bypass full decoder computation while retaining layer-local sliding-window attention. This nearly halves prefill computation, lowering the cost of processing new or uncached inputs in agentic workloads with growing contexts. CSA2 shares global KV across layers to reduce cache storage and reuses sparse selections to reduce indexing work. In the decoder, a Hierarchical Sparse Indexer restricts later indexers to a candidate pool selected by an earlier indexer, further reducing the number of entries scored per query.

We retain the shared and fine-grained routed experts of DeepSeekMoE~\citep{deepseekmoe}, and introduce modality-specific load balancing~\citep{noaux_tc} for image and text tokens. Single-Pass \mhc{}~\citep{mhc} revises residual-stream mixing to enable more efficient kernel fusion, and Engram~\citep{engram} adds sparsely accessed conditional memory. We omit the MTP module during backbone pre-training and use DSpark~\citep{cheng2026dspark} for speculative decoding. We train DSpark separately after the backbone pre-training stage. Additionally, we compress the main KV cache to FP4 to further reduce storage overhead. The following sections describe these components and the corresponding optimization changes.

\subsubsection{Multimodal Architecture}

The multimodal input pathway comprises a vision encoder and an MLP projector. For each input image, the vision encoder produces a spatial grid of visual features. A $3 \times 3$ pixel-unshuffle operation then rearranges each local neighborhood along the channel dimension, reducing the spatial resolution before the MLP projector maps the features to the hidden dimension of the language backbone. Finally, the resulting visual embeddings are inserted at the corresponding image-token positions in the input embedding sequence and processed jointly with text embeddings by the language backbone.

\paragraph{DeepSeek-ViT} 
We train a vision encoder named DeepSeek-ViT from scratch to natively process images at varying resolutions. We build DeepSeek-ViT on the Vision Transformer~\citep{vit} architecture with several modifications. To accommodate inputs of arbitrary resolutions, we replace standard absolute positional embeddings with 2D-RoPE. To align the ViT more closely with LLM design principles, we replace the patch embedding layer's convolution with a linear projection to ensure compatibility with the Muon optimizer. We also adopt RMSNorm~\citep{rmsnorm} for normalization and SwiGLU~\citep{shazeer2020glu} as the activation function. Before feeding visual features into the LLM, we apply a pixel-unshuffle operation with $3 \times 3$ downsampling to reduce the visual token count by a factor of nine, effectively supporting input resolutions up to approximately $1344 \times 1344$ pixels.

\paragraph{Multimodal Auxiliary-loss-free Load Balancing for MoEs}
Image and text tokens exhibit distinct representation distributions and may induce different expert-routing preferences in MoEs. Balancing their aggregate load may therefore obscure modality-specific imbalance. To address this issue, we extend auxiliary-loss-free load balancing~\citep{noaux_tc} by maintaining separate expert-wise correction biases for text and image tokens. During routing, each token uses the correction biases associated with its modality for expert selection, while the original routing scores are retained for weighting the selected expert outputs. After each training step, the two sets of biases are updated independently according to their respective expert loads. This design balances expert utilization within each modality and contributes to stable and efficient multimodal training.

\subsection{Causal Encoder-Decoder (CED)}
In agentic workflows, frequent tool calls generate extensive prefill requests, imposing severe computational overhead when KV caches miss. To alleviate this prefill bottleneck, we propose the Causal Encoder-Decoder (CED) architecture, inspired by YOCO~\citep{yoco}. YOCO reduces prefill computation by allowing the upper half of the layers to directly share the KV cache generated by the lower half. Building upon this concept, CED introduces a series of structural improvements to enhance both the overall KV cache capacity and the computational depth of KV generation. Consequently, CED successfully reduces nearly half of the prefill computation while maintaining performance comparable to the baseline.

For global attention, CED treats the bottom $L/2$ layers of the Transformer as the causal encoder. For the upper half layers (i.e., the decoder, $l > L/2$), the KV entries are not derived from their respective hidden states $H_l$. Instead, they are projected directly from the hidden state of the $(L/2)$-th layer, $H_{L/2}$, using layer-dependent projection weights ($W_l^{KV}$ and $W_l^{Z}$):
\begin{equation}
    C_l = H_{L/2} W_l^{KV}, \quad Z_l = H_{L/2} W_l^{Z}, \quad  l > \frac{L}{2},
\end{equation}
where $C$ and $Z$ represent the KV entries and their corresponding compression weights, respectively. This design allows CED to compute only the first half of the layers during the prefill phase, acquiring the upper-layer global KV cache with minimal computational cost.

For sliding window attention (SWA), CED maintains the conventional layer-wise computation across all layers. Specifically, for any layer $l$, the local keys and values are derived directly from the current layer's hidden state $H_l$.
This design effectively increases the computational depth of local KV generation. However, maintaining this layer-wise computation necessitates an SWA replay process. During the prefill phase, computing the SWA KV cache for the decoder requires processing an additional $n_{\mathrm{win}} \times L/2$ tokens (where $n_{\mathrm{win}}$ denotes the window size). For multi-turn interactions with short prompts per turn, this computational overhead in the decoder becomes non-negligible. Fortunately, prior work~\citep{powerattention} has shown that the actual effective receptive field of SWA is much smaller than the theoretical $n_{\mathrm{win}} \times L/2$. Motivated by this observation, we introduce Decoder SWA Bounded Replay, which only prefills the last $n_{\mathrm{win}}$ tokens of the prompt for the SWA computation, thereby significantly reducing the computational cost. Further details are provided in Section~\ref{sec:swa_bounded_replay}.

Overall, for a sequence length $N \gg n_{\mathrm{win}}$, CED reduces the prefill complexity from $\mathcal{O}(NL)$ to $\mathcal{O}(NL/2 + n_{\mathrm{win}}\times L/2) \approx \mathcal{O}(NL/2)$, effectively halving the overall computation. 

\begin{figure}[t]
\centering
\includegraphics[width=\textwidth]{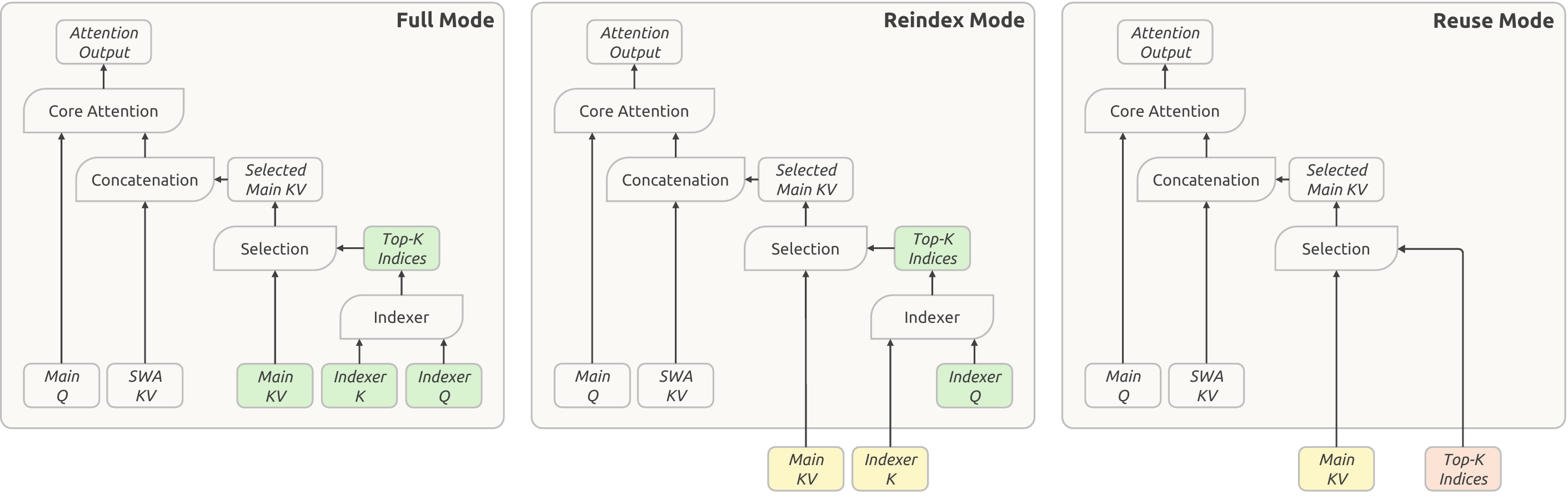}
\caption{
\textbf{Three operating modes of CSA2.}
The modes differ in how they obtain main KV, indexer K, and Top-K indices. Green blocks indicate quantities computed in the current layer; yellow blocks indicate main KV and indexer K reused from the most recent Full Mode layer; while red blocks indicate Top-K indices reused from the most recent index-producing (Full or Reindex Mode) layer. All three modes compute main Q and SWA KV in the current layer.
}
\label{fig:arch_csa2}
\end{figure}

\subsection{Compressed Sparse Attention 2 (CSA2)}
\label{sec:attention}

Serving long contexts requires controlling both KV cache storage and attention computation. These costs can be reduced along three multiplicative dimensions: the entry size, where GQA~\citep{ainslie2023gqa} reduces the number of KV heads and MLA~\citep{dsvii} shares a small latent across heads; the sequence dimension, where every $m$ tokens are compressed into one entry, like CSA and HCA in DeepSeek-V4~\citep{dsv4}; and the layer dimension, where some layers reuse the caches~\citep{brandon2024cla} and selections of other layers instead of keeping their own, or are replaced altogether by more efficient layers. Prior work has shown that compression along the layer dimension is effective: IndexCache~\citep{bai2026indexcache} reuses Top-K indices across layers to cut indexer computation; YOIO~\citep{sun2026yoio} computes the sparse routing once and shares it across all layers; and HySparse~\citep{gao2026hysparse} lets sparse layers reuse the KV cache of dense layers. However, index reuse alone saves no main KV storage, network-wide routing sharing limits performance, and hybrid designs still retain full attention layers; more importantly, none of these methods covers all three multiplicative dimensions.

CSA2 exploits the three dimensions jointly: it shares main KV and indexer K across layers and allows layers to reuse Top-K indices, with cache sharing and index reuse decoupled. It combines these reuse strategies with a simplified compressor and a Hierarchical Sparse Indexer that narrows the search domain of subsequent indexing layers in the Decoder.

Similar to CSA, CSA2 includes a lightweight indexer that scores the main KV entries using indexer Q and indexer K and selects the Top-K entries for each query. Each Q attends to the selected entries together with the layer-local sliding-window KV (SWA KV). CSA2 also includes the uncompressed main KV setting as a special case with a compression ratio of 1. Meanwhile, CSA2 simplifies both the compressor and the indexer. In CSA, a compression ratio of $m$ produces each main KV entry from $2m$ original KV cache entries, with overlapping source entries for adjacent compressed entries. It also includes absolute positional embedding to encode the positions of these $2m$ entries during compression. CSA2 removes this overlap and absolute positional embedding. In addition, CSA2 obtains indexer K by projecting main KV entries, replacing CSA's separate compression path from hidden states. Both designs simplify the implementation and increase the training efficiency.

Sections~\ref{sec:arch_csa2} and ~\ref{sec:arch_hsi} describe the cross-layer reuse strategies and the Hierarchical Sparse Indexer, respectively.

\subsubsection{Cross-Layer KV and Index Reuse}
\label{sec:arch_csa2}

Each CSA2 layer is statically assigned one of three modes: Full, Reindex, or Reuse. In all three modes, the layer computes its own query and SWA KV and uses them together with the selected main KV entries to produce a new attention output. The modes differ in how they obtain main KV, indexer K, and Top-K indices. Figure~\ref{fig:arch_csa2} illustrates the three modes.

\textbf{Full Mode.} The layer computes its own main KV and indexer Q, projects indexer K from that main KV, and runs the indexer to produce fresh Top-K indices. It therefore executes the complete CSA2 computation path and has the same component responsibilities as a complete CSA layer in DeepSeek-V4.

\textbf{Reindex Mode.} The layer reuses the most recent available \emph{main KV} from a preceding layer together with its corresponding \emph{indexer K}. The indexer computes its own query, rescores the reused keys, and produces fresh Top-K indices. This allows the sparse selection to change across layers while main KV and indexer K remain shared.

\textbf{Reuse Mode.} The layer reuses the most recent available \emph{main KV} and the latest \emph{Top-K indices} computed against that main KV by a preceding layer in Full or Reindex Mode. It performs attention using this selection without computing indexer Q or evaluating index scores.

Sharing main KV and indexer K reduces cache storage, while reusing Top-K indices avoids additional indexer computation. Reindex Mode preserves cache sharing while allowing the selected entries to change across layers.
When CSA2 is combined with CED, the decoder layer assigned to Full Mode computes its own global KV from the hidden state of the $(L/2)$-th layer, i.e. the last layer of the causal encoder. The Reindex and Reuse Modes are unchanged.

\begin{figure}[t]
\centering
\includegraphics[width=\textwidth]{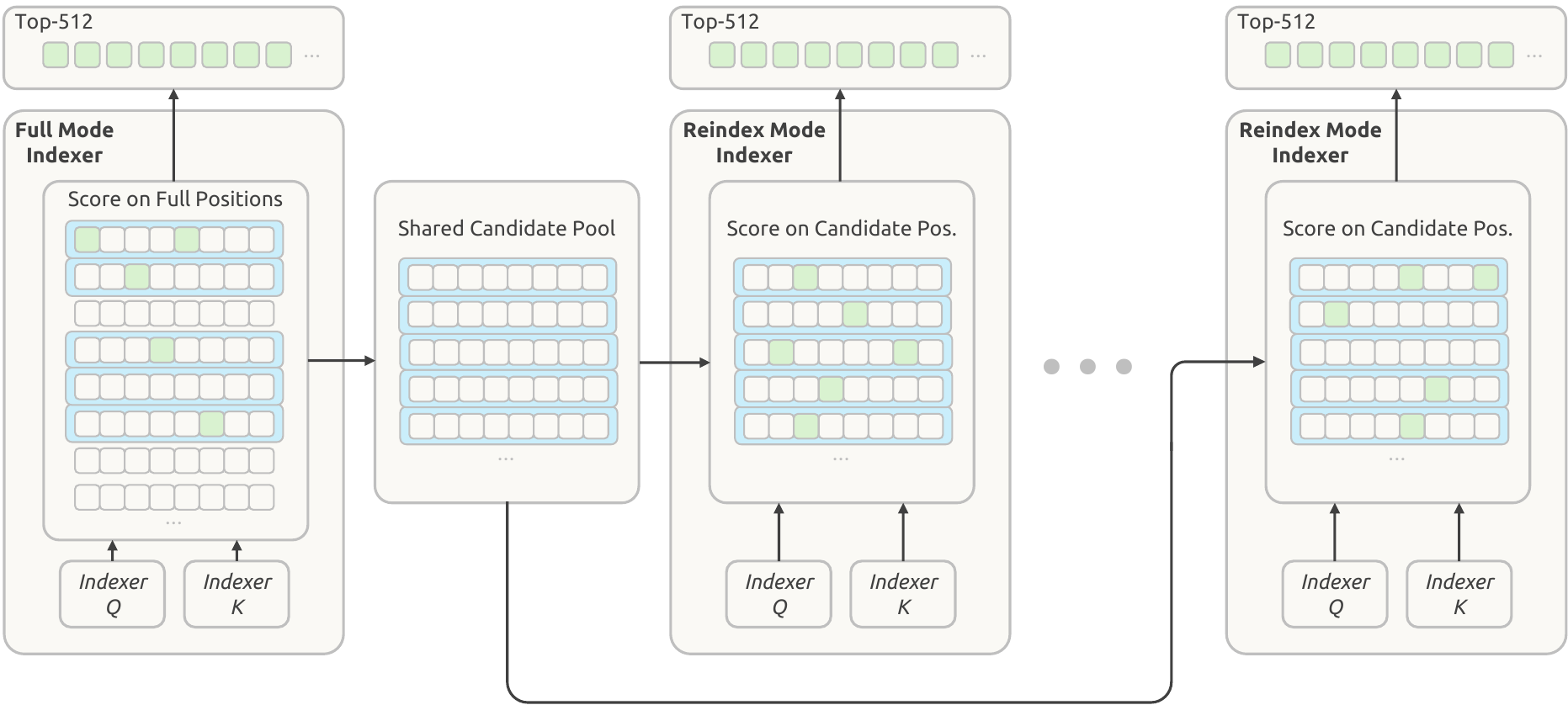}
\caption{
\textbf{Hierarchical Sparse Indexer.}
Each square represents a position; green squares mark selected indices, and blue rectangles mark blocks selected based on their maximum indexer scores. The decoder’s first CSA2 layer in Full mode selects its own Top-512 indices and builds a shared candidate pool from the selected blocks for subsequent layers. CSA2 layers in Reindex mode then select their Top-512 indices from this pool.
}
\label{fig:arch_hsi}
\end{figure}

\subsubsection{Hierarchical Sparse Indexer}
\label{sec:arch_hsi}

Cross-layer index reuse reduces the number of indexer evaluations, but the remaining indexers still score the full causally visible context. For extremely long contexts, this cost remains a major computational bottleneck. Prior work introduced indexer sparsity by scoring and pruning pooled block representations before token-level indexing~\citep{xu2026hisa}. We find that in the decoder, information from shallower indexers can naturally be used to restrict the candidates considered by deeper indexers without adding any extra state. We therefore introduce the Hierarchical Sparse Indexer, which is used only in the decoder of CED to reduce this repeated scoring during decode. For each query, the first layer assigned to Full Mode constructs a candidate pool that later re-indexing layers use as their search domain. For a fixed candidate-pool size, this changes the per-query cost of deeper indexers from linear in context length to constant. The mechanism is training-aware and introduced in post-training: the candidate restriction is applied identically during training and inference, so deeper indexers are optimized under the same search domain they use at inference. Figure~\ref{fig:arch_hsi} illustrates this process.

This first Full Mode layer scores all causally visible main KV positions and produces the Top-K indices for its own attention. It also performs blockwise candidate selection: each block is assigned the maximum index score among its positions, and the blocks with the highest scores are selected. It then collects the positions covered by the selected blocks into a candidate pool larger than the final Top-K set. For example, selecting 2,048 blocks with 8 positions each yields 16,384 candidate positions. This pool defines where later indexers search; the final Top-K selection determines which main KV entries each layer reads.

Subsequent layers in Reindex Mode score only the candidate positions for the corresponding query and select their own Top-K entries within that pool. Layers in Reuse Mode perform no new indexing and use the latest Top-K indices computed against the main KV they reuse. Thus, the candidate pool is shared across indexing layers, while their final selections can differ.

For a fixed candidate-pool size, the number of positions scored per query by each subsequent indexer is bounded independently of context length. The first Full Mode layer still scans the entire causally visible range. Hierarchical indexing therefore reduces the cost of later indexer evaluations while retaining the initial full-range pass.

\subsection{Efficient Architectural Extensions}

\subsubsection{Single-Pass \mhc{}}

In DeepSeek-V4, we introduced \mhc{}~\citep{mhc}, which maintains $n$ residual streams between adjacent Transformer blocks. For each token, we denote these streams by $X_l\in\mathbb{R}^{n\times d}$, where $l$ is the block index and $d$ is the hidden dimension. The streams are updated as follows:
\begin{equation}
    \label{eq:mhc}
    X_{l+1}=B_lX_l+C_l\,\mathcal{F}_l(A_lX_l),\qquad (A_l,B_l,C_l)=\mathcal{H}(X_l),
\end{equation}
where $A_l\in\mathbb{R}^{1\times n}$, $C_l\in\mathbb{R}^{n\times1}$ and $B_l\in\mathbb{R}^{n\times n}$ are token-wise coefficients predicted from $X_l$. The coefficient predictor $\mathcal{H}$ includes normalization and projection.

Ideally, the residual transformation between two blocks is a single map from $(X_{l-1}, Y_{l-1})$ to $(X_l, \hat{X}_l)$, where $\hat{X}_l=A_lX_l$ is the current block input and $Y_{l-1}=\mathcal{F}_{l-1}(\hat{X}_{l-1})$ is the previous block output. Such a map requires $(n+1)d$ reads and $(n+1)d$ writes, giving a lower bound of $(2n+2)d$ on activation memory traffic.
In practice, DeepSeek-V4 uses a multi-pass implementation of \eqref{eq:mhc}, with three kernels that execute sequentially due to data dependencies:
\begin{align}
    X_l &= B_{l-1}X_{l-1}+C_{l-1}Y_{l-1} && \text{Residual update, contraction over $n$}\\
    (A_l,B_l,C_l) &= \mathcal{H}(X_l)     && \text{Coefficients, contraction over $nd$}\\
    \hat{X}_l &= A_lX_l                   && \text{Input mixing, contraction over $n$}
\end{align}
The three kernels read $(n+1)d$, $nd$ and $nd$ values respectively and write $(n+1)d$ in total. Including the pre-norm in $\mathcal{F}_l$, the total activation memory traffic is $(4n+4)d$, twice the lower bound.

In $\mathcal{H}$, the normalization weights are folded into the projection weights offline, and the RMS division is applied after the projection. Two of the three stages can therefore share one traversal of the residual: the residual update does not require a reduction across the hidden dimension, so each tile of $X_l$ can be computed and immediately used to accumulate the projection outputs and the sum of squares needed to compute the RMS. 
Input mixing cannot be fused into this pass because $A_l$ is not available until the reduction over all hidden tiles is complete. It therefore requires a second read of $X_l$. This second pass can also incorporate input pre-norm. This two-pass implementation would require $(3n+2)d$ activation reads and writes in total, one additional read of $X_l$ compared with the lower bound.

We therefore introduce \textbf{Single-Pass \mhc{}}, which shifts the input-mixing coefficients by one block, i.e. every block consumes the mixing coefficients produced by the previous one, so that the above dependency disappears:
\begin{equation}
    X_{l+1}=B_lX_l+C_l\,\mathcal{F}_l(A_{l-1}X_l),\qquad (A_l,B_l,C_l)=\mathcal{H}(X_l).
\end{equation}

Input mixing now uses $A_{l-1}$ instead of $A_l$, so it no longer depends on the coefficients computed from $X_l$. Each tile of $X_l$ can therefore be used immediately for both input mixing and coefficient prediction, without waiting for the full reduction. Empirically, this shift incurs negligible performance degradation.

For pre-training, we keep the existing multi-kernel implementation, since the shift only changes which mixing coefficients each block applies. 
For deployment, we fuse residual update, input mixing, and coefficient prediction into a single kernel, \textbf{Mega-\mhc{}}. The kernel implements \mhc{} with $(3n+2)d$ activation reads and writes and Single-Pass \mhc{} with $(2n+2)d$ activation reads and writes. 
Mega-\mhc{} processes $X_l$ in tiles along the hidden dimension. Each tile is used to compute the mixed input and to accumulate the quantities needed to predict $(A_l,B_l,C_l)$ for the next block. The kernel also incorporates input pre-norm and FP8 conversion.
The residual is thereby read once and written once, attaining the $(n+1)d$ reads and $(n+1)d$ writes of the ideal map and halving the activation memory traffic of our original implementation.

\subsubsection{Engram}

We augment DeepSeek-V4.1-Flash with Engram~\citep{cheng2026conditional}, the conditional memory module introduced in our previous work to decouple memorization from computation. We follow the original Engram design---tokenizer compression, multi-head hashing, context-aware gating, and multi-branch integration---with two modifications. First, we omit the short causal convolution because its performance gains do not justify the added complexity in our inference stack. Second, we optimize the Engram embedding with momentum-based update followed by Sinkhorn balancing, as detailed in Section~\ref{sec:optimization}.

We allocate 196B Engram parameters evenly across two modules. Each module uses $N$-gram orders $\{2,3,4\}$, with 8 hash heads and a total embedding dimension of 2048 per order. Each head indexes a table of approximately 16M entries, with table sizes chosen to be distinct primes. Both the embedding tables and the key/value projections use FP8 precision. The modules are placed at layers 1 and 14 (zero-indexed) to balance memory usage across training pipeline stages. During inference, deterministic addressing enables embeddings to be prefetched from host memory via background RDMA transfers, with prefetching for the first module overlapping computation in the first Transformer block. Further implementation details for Engram training and inference are discussed in Section~\ref{sec:engram-infrastructure}.

\subsubsection{DSpark}

We equip DeepSeek-V4.1-Flash with DSpark~\citep{cheng2026dspark}, a speculative decoding module that combines semi-autoregressive drafting with confidence-scheduled verification. The drafter comprises three Transformer blocks with a sliding attention window of 128 tokens. A single forward pass through these blocks computes base logits for five draft positions in parallel, while a lightweight Markov head models dependencies among the draft tokens. A confidence head predicts per-position conditional acceptance probabilities, which are used to estimate prefix survival probabilities. The scheduler combines these estimates with profiled engine throughput curves to dynamically select the verification length for each request, aiming to maximize expected system-wide token throughput under the current system load.

Unlike the MTP module in DeepSeek-V3~\citep{dsv3}, which is trained jointly with the backbone throughout pre-training, DSpark is introduced in a dedicated stage after pre-training. In this stage, we train only DSpark while keeping the backbone frozen. During post-training, we continue to train DSpark alongside the backbone, without propagating gradients from the DSpark objective into the backbone. This keeps DSpark aligned with the evolving policy, enabling it to accelerate both online serving and rollout generation for RL and OPD.

\subsubsection{FP4 Main KV Cache}

Long-context agent workloads require large per-request KV caches, increasing serving costs. DeepSeek-V4 already uses quantization-aware training (QAT)~\citep{QAT} for FP4 indexer queries and keys, accelerating index computation and reducing the indexer cache size. We adopt the OCP-standard MXFP4 format~\citep{OCP_MXFormat} to support as many hardware platforms as possible, despite the higher accuracy of alternative formats in our experiments.
We now extend QAT to the main KV cache, where FP4 reduces storage rather than accelerates matrix multiplication. Dequantizing cached values before attention allows us to use a more accurate format without requiring native matrix-multiplication support for that format, preserving compatibility across hardware platforms.

Among the approximately four-bit formats evaluated, we select E2M1 with one E4M3 scale per 16 channels, following NVFP4~\citep{NVFP4} but omitting its second-level global scale to balance accuracy and simplicity.
Omitting this scale leaves ample dynamic range for the main KV cache: the format supports magnitudes up to $448\times6=2688$, far above the cache's magnitude bound. In DeepSeek-V4.1-Flash, the largest trained RMSNorm weight magnitude is approximately $1$. After RMS normalization, the L2 norm of the $512$-channel KV latent is at most approximately $\sqrt{512}$. RoPE preserves this norm, so the maximum absolute value across channels after rotation is also bounded by approximately $\sqrt{512}\approx22.6$. Besides, the maximum magnitude observed during training is around $10$. Therefore, omitting the global scale causes no measurable decrease in accuracy and simplifies the cache layout.

To enable FP4 main KV cache storage in DeepSeek-V4.1-Flash, we introduce QAT during post-training. The non-RoPE and RoPE components use the same quantization format. We quantize the cache after RoPE: quantizing before RoPE yields only a marginal accuracy improvement in our experiments and would introduce additional overhead during decoding.
We retain FP8 for the SWA KV cache due to its sensitivity to quantization.
Compared with the FP8 main KV cache in DeepSeek-V4, this format nearly halves the storage footprint, both in HBM and when offloaded to SSD.

\subsection{Optimization}
\label{sec:optimization}

Building upon the optimization configuration used in DeepSeek-V4, we make some new modifications to better align with the architectural design. 

First, we use head-wise Muon, where Query weights are split by head before applying the Muon update.  Here, we briefly discuss the motivation for such a design. 
By viewing Muon as a preconditioned gradient descent, vanilla Muon uses one preconditioner for all heads, whereas head-wise Muon provides different preconditioners for different heads. This design can better handle the heterogeneity across attention heads~\citetext{\citealp{zhang2024transformers}; \citealp[Section 3]
  {zhang2026principles}}.
As a result,  we observe that head-wise Muon outperforms vanilla Muon. The empirical advantage of head-wise Muon is also validated in GLM 5 \citep{zeng2026glm} and Kimi-K3 \citep{kimi_k3}.

Second, applying Adam to the newly introduced Engram parameters substantially increases the optimizer-state memory footprint. 
To reduce memory usage during training, we instead optimize the Engram embedding tables, token embedding, and prediction head using a momentum-based update followed by Sinkhorn balancing. 
Sinkhorn balancing has previously been applied to linear-layer weight matrices in SinkGD~\citep{sinkgd}; here, we extend it to these large parameter matrices.
Like Muon, this approach requires only a momentum buffer while empirically outperforming Adam.

\paragraph{Basic Configurations.}
We retain AdamW~\citep{adamW} for normalization-layer weights and other non-matrix parameters, including biases and scaling factors. 
We use Muon~\citep{muon} for the weight matrices of linear transformations in the language-model backbone, the Engram projection layers, and the vision-language projector. We use head-wise Muon for Query and Key weights.
We apply decoupled weight decay and Nesterov momentum to Muon~\citep{nesterov,muon_kimi}; normalization-layer weights are also subject to weight decay, whereas biases and scaling factors are not. 
The Sinkhorn-balanced update also uses Nesterov momentum but does not apply weight decay. 
During pre-training, we keep the vision encoder frozen until the learning-rate decay stage, while its final normalization layer and the vision--language projector remain trainable. At the onset of learning-rate decay, we unfreeze the vision encoder and optimize it jointly with the LLM with a smaller learning rate.

\paragraph{Sinkhorn-Balanced Updates for Engram / Embedding / Prediction Head.}
The complete procedure is summarized in Algorithm~\ref{alg:sinkhorn_momentum}. 
At a high level, it follows the same workflow as Muon, with Sinkhorn balancing taking the place of Newton--Schulz orthogonalization.
We denote the larger matrix dimension by $m$, which corresponds to the vocabulary size for embedding tables and prediction heads, and denote the hidden dimension by $n$.

\begin{algorithm}[t]
\caption{Momentum update with Sinkhorn balancing}
\label{alg:sinkhorn_momentum}
\begin{algorithmic}[1]
\Require Weight $W_t \in \mathbb{R}^{m \times n}$, gradient $G_t$, momentum $M_{t-1}$, momentum coefficient $\beta$, base learning rate $\eta_t$, learning-rate correction $\gamma$, numerical constants $\epsilon$ and $\tau$, and an odd number of normalization steps $K$
\State $M_t \gets \beta M_{t-1} + (1-\beta)G_t$
\State $\widehat{G}_t \gets \beta M_t + (1-\beta)G_t$ \Comment{Nesterov momentum}
\State $\rho_i \gets \lVert \widehat{G}_{t,i,:}\rVert_2$ and $\bar{\rho} \gets \frac{1}{m}\sum_{i=1}^{m}\rho_i$
\State $U^{(0)} \gets \widehat{G}_t$
\State Set $U^{(0)}_{i,:} \gets 0$ if $\rho_i \leq \tau\bar{\rho}$ \Comment{Mask near-zero rows}
\For{$k=1,\ldots,K$}
    \If{$k$ is odd}
        \ForAll{$i=1,\ldots,m$}
            \State $U^{(k)}_{i,:} \gets U^{(k-1)}_{i,:}/(\lVert U^{(k-1)}_{i,:}\rVert_2+\epsilon)$
        \EndFor
    \Else
        \ForAll{$j=1,\ldots,n$}
            \State $U^{(k)}_{:,j} \gets U^{(k-1)}_{:,j}/(\lVert U^{(k-1)}_{:,j}\rVert_2+\epsilon)$
        \EndFor
    \EndIf
\EndFor
\State $\Delta_t \gets \sqrt{n}\,U^{(K)}$ \Comment{Convert unit row $\ell_2$ norm to unit row RMS}
\State $\widetilde{\eta}_t \gets \gamma\eta_t$ \Comment{Match the update magnitude of Adam}
\State $W_{t+1} \gets W_t - \widetilde{\eta}_t\Delta_t$
\end{algorithmic}
\end{algorithm}

Given the Nesterov momentum update $\widehat{G}_t$, Sinkhorn balancing finds diagonal scaling matrices $D_r$ and $D_c$ such that
\begin{equation}
\Delta_t = \sqrt{n}\,U^{(K)} = \sqrt{n}\,D_r\widehat{G}_tD_c,
\qquad
\frac{1}{n}\sum_{j=1}^{n}(\Delta_t)_{ij}^{2} \approx 1,
\qquad
\frac{1}{m}\sum_{i=1}^{m}(\Delta_t)_{ij}^{2} \approx 1,
\label{eq:sinkhorn_balance}
\end{equation}
Thus, the procedure approximately equalizes the row-wise and column-wise RMS of the update matrix. Here, one row corresponds to one token index or n-gram identity; and one column encodes one hidden feature. Sinkhorn balancing exploits this token--feature structure by normalizing along both rows and columns.
For numerical stability, rows satisfying $\rho_i \leq \tau\bar{\rho}$ are masked. The factor $\sqrt{n}$ converts unit row $\ell_2$ norm into unit row-wise RMS. Separately, we adjust the effective learning rate as $\widetilde{\eta}_t=\gamma\eta_t$ to match the update magnitude of Adam. We set $\gamma=0.18$, which is close to the factor $0.2$ used in Moonlight~\citep{muon_kimi}.

More broadly, Sinkhorn balancing is closely related to optimizers that exploit  matrix or tensor axis structure~\citep{adafactor,adam-mini,wen2025sron,glentis2025memory, deng2026rmnp,yuan2026nora,xu2026width}. For  example, Adafactor~\citep{adafactor} conducts row- and column-wise normalization in a different manner, and Adam-mini~\citep{adam-mini} uses an alternative row-wise normalization for embedding tables and prediction head. These normalization strategies may differ in optimization performance and communication overhead. We leave more detailed investigation as a future direction.

\section{General Infrastructures}
\label{sec:infra}

\subsection{Training Infrastructure}
\subsubsection{Multimodal Training Infrastructure}
\label{sec:infra-multimodal}

\paragraph{Communication-Computation Overlap in Contrastive Learning.}
The vision encoder is first optimized with a contrastive objective before being fine-tuned with a generative next-token prediction loss. In the contrastive phase, the loss is computed over a full batch of text and vision pairs, so the features of both modalities must be all-gathered across data-parallel ranks, incurring substantial communication. Because the gradient of the text features depends only on the gathered visual features---and, symmetrically, the gradient of the visual features depends only on the gathered text features---each all-gather can be overlapped with the forward or backward pass instead of stalling the pipeline:
\begin{multline*}
\text{Forward}(V)
\;\rightarrow\; \bigl(\text{Forward}(T) \;\|\; \text{AllGather}(V)\bigr)
\;\rightarrow\; \nabla_{\text{Text}}\\
\;\rightarrow\; \bigl(\text{Backward}(T) \;\|\; \text{AllGather}(T)\bigr)
\;\rightarrow\; \nabla_{\text{Vision}}
\;\rightarrow\; \text{Backward}(V),
\end{multline*}
where $V$ and $T$ denote the visual and text features, $(A\,\|\,C)$ denotes the overlap of computation $A$ with communication $C$, and $\nabla$ denotes the gradient computation. In this schedule, the visual features are gathered during the text forward pass and the text features during the text backward pass, so that both all-gathers are hidden entirely behind useful computation.

\paragraph{End-to-End Parallelism.}
To handle the model and data heterogeneity between the vision encoder and the LLM \citep{zhang2025disttrain}, we adopt the disaggregated encoder design used in recent training systems~\citep{longcat-flash-omni,K2.5}. The vision encoder is replicated outside the LLM parameter tree, and each training step is divided into three phases: vision encoder forward, LLM forward/backward, and vision encoder backward. This separation prevents interference between vision encoder and LLM computation. Load-balanced vision processing is confined to the first and last phases, while the LLM phase remains free of vision computation and preserves the parallel strategy of text-only training.

\paragraph{Long-Sequence Multimodal Training Optimization.}
DeepSeek-V4.1-Flash is trained on sequences of up to one million tokens, with a substantial share of training occurring at ultra-long sequence lengths in both the pre-training and post-training stages. At these sequence lengths, multimodal samples create heavy I/O, CPU, and memory bottlenecks.

\begin{itemize}
    \item \textbf{Balanced image sharding.} During pre-training, a single ultra-long, image-dense sequence can exhaust one host's I/O, CPU, and memory during loading, so the images of each sequence are sharded across the CP ranks with load balancing, and each image is loaded exactly once. With images read once, loading stays hidden behind compute whenever
\[
\frac{N\times \rho}{B_{\mathrm{IO}}} < \frac{N\times C}{B_{\mathrm{GPU}}}
\quad\Longleftrightarrow\quad
\rho < \frac{B_{\mathrm{IO}}}{B_{\mathrm{GPU}}}\,C,
\]
where $N$ is the token count, $\rho$ the raw bytes per token, $C$ the per-token compute, and $B_{\mathrm{IO}}$, $B_{\mathrm{GPU}}$ the file-system and GPU bandwidths. Since $N$ cancels, the criterion involves only per-token quantities ($\rho$ and $C$), independent of sequence length and cluster size; $\rho$ is set by the vision-module configuration (e.g., the resolution cap or the spatial downsample). Storage throughput therefore becomes a bottleneck only for small models with low per-token compute, as in ablations, while production-scale models remain compute-bound.

    \item \textbf{Incremental image transfer.} Besides the balanced sharding above, the reinforcement-learning rollout transfers images to the inference engine only incrementally, and caches the engine's CPU-side decoding and preprocessing outputs on a distributed file system for reuse across rollouts and subsequent training.
\end{itemize}

\subsubsection{Attention Sharing Training for CSA2}
In Section~\ref{sec:attention}, we introduce CSA2, an attention sharing method with three modes, some of which involve sharing one or more of the main KV, the indexer K, and the Top-K indices across multiple layers. Supporting CSA2 in large-scale distributed training requires additional coordination beyond the attention computation itself. In particular, layers that share attention components may be placed on different pipeline stages, making direct module reuse incompatible with conventional stage-local execution. Therefore, we adopt several designs to support CSA2 training.

\textbf{Shadow indexers} address this issue by placing a lightweight executable replica on each participating stage while retaining a single logical owner for the shared parameters. The owner remains responsible for optimization and checkpointing, whereas parameter synchronization and gradient aggregation keep the shadow replicas consistent throughout training. This design preserves the original model semantics without requiring the pipeline scheduler to treat shared layers as a special execution unit. 

\textbf{Pipeline payload extensions} provide the intermediate representations and sparse routing information required by downstream consumers when the source and consumer layers cross a pipeline boundary. These states are incorporated into the existing point-to-point communication path and partitioned consistently with context parallelism, avoiding unnecessary replication while maintaining the corresponding gradient flow. 

\textbf{Micro-batch-level shared-state management} tracks the states associated with concurrently active pipeline micro-batches and coordinates their lifetimes across forward execution, activation recomputation, and backward propagation. Shared states are retained until their final consumer has completed and are then released promptly to limit additional memory overhead. The same runtime abstraction also handles stage placement and source--consumer relationships, allowing the attention implementation to access shared states without depending on the physical pipeline layout. 

Together with lightweight adaptations to the optimizer, checkpointing, warm-up, and computation-graph tracing workflows, these mechanisms enable CSA2 to operate transparently under the existing distributed training interface and pipeline schedules.

\subsubsection{Engram}
\label{sec:engram-infrastructure}
Engram embedding tables are partitioned by row across dedicated process groups of engram parallel size.
The group size controls the trade-off between per-device memory usage and the communication scope of embedding lookups.
Optimizer states are further sharded across replicas of each table partition.
Engram lookup indices depend solely on the input token sequence.
Embedding prefetch is therefore initiated for the entire local batch before each pipeline stage begins processing microbatches for the current training step, minimizing interference with pipeline execution.
Embedding gradients are buffered during backward and returned to their owning ranks after the backbone backward pass.
For efficient integration with multimodal training, embedding prefetch and gradient transfers are scheduled to overlap with the vision encoder's forward and backward computation.
Embeddings are stored and fetched in FP8, with the retrieved values and scaling factors passed directly to the following GEMM. For Engram table updates, Sinkhorn normalization maintains row and column scaling vectors across iterations to avoid repeated writes of the full normalized matrix. Row normalization and the accumulation of partial column statistics are fused into a single kernel to further reduce memory traffic. During RL rollouts, Engram embedding tables remain resident in GPU memory.
This placement reduces host memory pressure and helps avoid out-of-memory failures caused by host memory fragmentation.

\subsection{Inference System}

DeepSeek-V4.1-Flash is designed with inference efficiency as a first-class concern. Although its architecture is conceptually complex, the resulting inference kernel flow is remarkably concise. Through reasonable kernel fusion, we encapsulate the intricate operations and keep hardware resources fully pipelined inside a small number of fused kernels---including the fused-RoPE-attention-RoPE-cast kernel in FlashMLA~\citep{flashmla2025}, the Mega-Gate, Mega-\mhc, and Mega-MoE kernels in DeepGEMM~\citep{deepgemm2025}, the kernels in TileKernels~\citep{tilekernels}, and the TopK kernel in DeepSelect~\citep{deepselect2026}. As a result, the vast majority of Transformer layers---those whose CSA2 operates in Reuse Mode---execute with only 15 kernels during prefill and 11 during decode, thereby achieving both high-throughput and low-latency inference.

At the deployment level, we adopt Encoder--Prefill--Decode (EPD) disaggregation, enabling vision encoding, prefill, and decoding to scale independently and overlap in execution.

\subsubsection{Persistent KV Cache Management}
\label{sec:persistent_kvcache_management}
Under identical workloads, V4.1 reduces the persistent KV cache footprint to 1/8 of that of V4. Two multiplicative factors account for this reduction: the persistent KV cache no longer stores SWA KV, which almost halves its size, and the global KV retained in it is further compressed to 1/4 of V4's footprint through architectural and precision optimizations.

In the V4 deployment, SWA KV accounts for nearly half of the persistent KV cache capacity. Within this cache, global KV and SWA KV are managed independently, governed by an LRU eviction policy. Global KV is stored in its entirety, and upon a hit, the complete prefix is reused. In contrast, SWA KV is cached at two specific points---the end of the prompt and the end of the output---to facilitate regeneration and multi-turn sessions; a hit allows computation to resume from that cached position. To maintain a high hit rate, we configured a sufficiently large persistent KV cache on SSD such that, under typical workloads, both types of KV remain resident for over 72 hours.
Despite retaining only $n_\mathrm{win}$ KV entries at designated positions, the uncompressed SWA KV cache still incurs a substantial storage overhead, especially in multi‑turn conversations with short turns.

Persistently storing SWA KV is both costly and ineffective, because its access pattern does not match the persistent KV cache's long retention policy. Unlike global KV, which exhibits long-tail reuse, SWA KV is reused only within a narrow, minute-scale window inside an active session and becomes dead once the session ends or the next turn begins. The V4 technical report proposed \textit{Zero SWA Caching}, which avoids the storage overhead by recomputing missing SWA KV. Exact recovery, however, requires a full forward pass over \( L \times n_{\mathrm{win}} \) tokens, whose cost proved prohibitive in production deployments.

V4.1 therefore revises persistent KV cache management as follows:

\begin{enumerate}
    \item SWA KV is no longer cached in the persistent KV cache and instead stored in a distributed memory pool provisioned from 10\% of the host DRAM on each machine. Although this pool is far smaller in aggregate capacity, its short TTL (only minutes) allows expired entries to be recycled immediately for new sessions; under real-world workloads, this high turnover suffices to serve the vast majority of concurrent active sessions. Global KV remains in the persistent KV cache with a guaranteed lifetime of at least 72 hours.

    \item Evicting SWA KV inevitably causes misses, which stay affordable thanks to a lightweight fallback, \textit{Encoder SWA Bounded Replay} (detailed in Section \ref{sec:swa_bounded_replay}). For the inevitable but infrequent requests that hit global KV but miss SWA KV, it recovers the missing state by recomputing only \( n_{\mathrm{win}} \) tokens instead of a full \( L \times n_{\mathrm{win}} \)-token forward pass. This bounded replay is the cornerstone of the design: it turns a catastrophic miss into a graceful, inexpensive degradation, thereby justifying the removal of SWA KV from the persistent KV cache.
\end{enumerate}

\subsubsection{SWA Bounded Replay}
\label{sec:swa_bounded_replay}
Since SWA dependencies accumulate across layers, exactly reconstructing the SWA KV of $L$ layers would require replaying $L \times n_{\mathrm{win}}$ tokens. SWA Bounded Replay instead replays only the most recent $n_{\mathrm{win}}$ tokens and truncates SWA to the replay segment, accepting approximate states: for a replay starting at position $s$, a query at position $i$ attends to SWA keys in $\bigl[\max(s,\,i-W+1),\,i\bigr]$.

\paragraph{Encoder SWA Bounded Replay.}
Encoder SWA Bounded Replay makes prefix caching depend only on global KV, allowing SWA KV to be removed from the persistent KV cache.

When the encoder SWA KV is missing, we replay the last $n_{\mathrm{win}}$ tokens of the cached prefix and process them together with the uncached suffix. The replayed tokens regenerate only SWA KV, reusing the cached global KV without recomputation or overwriting, while the uncached suffix generates both global KV and SWA KV.

By design, the replayed prefix state is approximate, so the global KV and SWA KV computed for the uncached suffix depend on the cache-hit position and are not mathematically identical across positions.
Encouragingly, our experimental evidence confirms that this bounded replay strategy barely compromises response quality.

\paragraph{Decoder SWA Bounded Replay.}
Decoder SWA Bounded Replay bounds the decoder forward pass to $n_{\mathrm{win}}$ tokens, nearly halving total prefill computation.

Under CED, decoder global KV is projected from the final encoder hidden states. The only obstacle to ending prefill at the encoder is decoder SWA KV, which is generated from each decoder layer's own hidden states and is needed by the first decode steps. Since we never cache decoder SWA KV, exactly reconstructing it requires running the $\frac{L}{2}$ decoder layers over the last $\frac{L}{2} \times n_{\mathrm{win}}$ prompt tokens, which is expensive when a short uncached suffix follows a long cached prefix. Therefore, we also apply the bounded replay strategy to this scenario: at every prefill, we replay the last $n_{\mathrm{win}}$ tokens of the prompt, feed their encoder outputs through the decoder layers under the same SWA truncation, and use the resulting decoder SWA KV only for decoding, not for prefix caching.

By design, the reconstructed decoder SWA KV is not mathematically equivalent to that from a full decoder forward pass. Also, we find that this strategy has only a negligible impact on response quality. For added safety, we additionally simulate the same replay during post-training for train-aware adaptation.

\section{Pre-Training}
\label{sec:pre-training}

\subsection{Data Construction}
\paragraph{Text Data Curation}
In pursuit of higher intelligence,  we go beyond the general, sample-level quality reflected by small-scale data experiments and focus more on the holistic interactions among diverse corpora that offer unique information gains.
We adopt a more systematic and standardized data construction pipeline to improve data quality and optimize the data mixture.
Specifically, based on more comprehensive evaluations, a scaling ladder over model parameters and training data is carefully designed to guide large-scale training runs.
We filter out model-generated content with limited information gain, including outputs from less capable models and low-quality machine-translated text. We regard such content as implicit duplication, as it largely reformulates existing information and may become detrimental over long training horizons.
We also explore model-in-the-loop data iteration approaches as a foundation for future large-scale synthetic data. 
In addition, we involve more domain experts to construct fine-grained data quality evaluation dimensions. 
Compared with the previous version, the new corpus incorporates more recent code from newly released open-source repositories, commits, libraries, and emerging frameworks to cover a broader range of programming languages and better reflects contemporary real-world software engineering scenarios.

\paragraph{Multimodal Data Curation}
Our multimodal pre-training dataset primarily comprises three types of data: image-text pairs, interleaved image-text data, and domain-specific data. 
Operating on the premise that raw web data naturally provides rich multimodal knowledge, we refrained from large-scale data synthesis; instead, we prioritized cleaning and utilizing the data in its native form to achieve the most direct and scalable visual knowledge compression during pre-training. During the initial data collection, we found that our crawling system was overly biased toward text-centric web content; we therefore re-bootstrapped it from Common Crawl to improve its coverage of multimodal sources.
For image-text data, we extract images together with their associated alt text from webpages, filter them by applying an image-text relevance threshold, and deduplicate them based on image semantics. For interleaved data, we build this subset predominantly from webpages and PDFs. 
Processing large-scale multimodal corpora usually incurs higher CPU and disk storage costs than processing text-only corpora, which motivated us to organize the interleaved-data construction into progressively more expensive stages.
Before image retrieval, we apply heuristic and statistical filtering, deduplication, and quality models to select high-value documents.
The surviving documents are then assembled into interleaved image–text sequences, where filtering and deduplication are applied again in an image-aware manner.
Finally, we employ SmolVLM~\citep{marafioti2025smolvlm} to conduct strict quality scoring on the image-text content, thereby extracting high-quality interleaved data.
Documents filtered out during this process are partly recycled into additional image-text pairs via screening and recombination.
To compensate for the inherent limitations of web-gathered data, we also incorporate domain-specific datasets to boost the model's capabilities in fine-grained visual perception (e.g., visual grounding and pointing), optical character recognition (OCR), and the acquisition of long-tail knowledge. We also collect extensive image-code pairs and computer-use trajectories to improve multimodal agentic understanding.

\paragraph{Data Integration and Deduplication}
As our text-only and multimodal data were processed through distinct pipelines, we constructed the final training corpus as the union of both data sources. For overlapping samples, we replace the text-only versions with their multimodal counterparts and use the larger epoch count of the two configurations. After this substitution, the resulting corpus uses a $7{:}1$ token ratio of text-only to multimodal data.
We minimize sample overlap during pre-training and context extension by jointly prefetching and assigning training samples.
Ultra-long documents are deterministically pre-split before mixing to ensure a uniform distribution of training tokens across data shards and training steps. We further enhance our
best-fit packing algorithm, achieving a padding rate of at most $10^{-4}$.

\subsection{Pre-Training Setups}

\subsubsection{Model Setups}

We set the number of Transformer layers to 40 and the hidden dimension $d$ to 5120.
We adopt a Causal Encoder-Decoder architecture, with 20 layers in the encoder and 20 layers in the decoder.
For the first two layers, we use pure sliding window attention.
The remaining 18 encoder layers use CSA2 with a compression rate of $m=2$.
These layers are divided into three identically configured groups of six layers.
In each group, the first layer operates in Full Mode, and the remaining five layers operate in Reuse Mode.
The 20 decoder layers use CSA2 with a compression rate of $m=1$.
These layers are divided into five groups of four layers.
In the first group, the first layer operates in Full Mode, and the remaining three layers operate in Reuse Mode.
The remaining four groups share the same configuration: the first layer operates in Reindex Mode, and the remaining three layers operate in Reuse Mode.
For all CSA2 layers, we set the number of indexer query heads to 32, the indexer head dimension to 128, and the number of KV entries selected for sparse attention (i.e., attention top-k) to 512.
We set the number of query heads to 64, the head dimension to 512, and the query compression dimension to 1280.
For the Hierarchical Sparse Indexer, we select a maximum of 2,048 blocks with 8 positions, yielding up to 16,384 candidate positions in total.
The number of output projection groups is set to 8, and the dimension of each intermediate attention output is set to 1024.
For the additional branch of sliding window attention, the window size $n_{\mathrm{win}}$ is set to 128.
We employ MoE layers in all Transformer blocks, using SwiGLU activation function with clamping~\citep{gpt_oss} at a threshold of 10.
Each MoE layer consists of 1 shared expert and 384 routed experts, where the intermediate hidden dimension of each expert is 2304.
Among the routed experts, 6 experts will be activated for each token.
As for \mhc{}, the expansion factor is set to 4, and the number of Sinkhorn-Knopp iterations is set to 20.
For the vision encoder, we set its number of layers to 32, the hidden dimension to 1024, the number of attention heads to 16, and the image patch size to 14. The vision MLP projector has 2 layers with a hidden dimension of 5120.
Under this configuration, DeepSeek-V4.1-Flash comprises 552B backbone parameters, with 8B activated per token during prefill and 16B during decode. 

\subsubsection{Training Setups}

We employ the Muon optimizer~\citep{muon,muon_kimi} for the parameters of linear transformations, use AdamW optimizer~\citep{adamW} for the weights of all RMSNorm modules and other non-matrix parameters, and use Sinkhorn-balanced update for all embeddings and prediction head.
For AdamW, we set its hyper-parameters to $\beta_1=0.9$, $\beta_2=0.95$, $\epsilon=10^{-20}$, and $\mathrm{weight\_decay}=0.1$.
For Muon, we set the momentum to $0.95$ and the weight decay to $0.1$, and rescale the RMS of each update matrix to $0.18$ for reutilization of the AdamW learning rate.
For Sinkhorn-balanced updating, we use the same momentum coefficient and learning-rate correction factor as for Muon and set $K=11$, $\tau=10^{-3}$, $\epsilon=10^{-20}$.
Following~\citep{engram}, the learning rate of Engram is scaled by $5\times$.
We train DeepSeek-V4.1-Flash on 45T tokens of multimodal data with no instability.
We keep the batch size fixed at 100.6 million tokens throughout training.
The learning rate is linearly warmed up over the first 2000 steps and then maintained at $2.6 \times 10^{-4}$ until 28T tokens.
Between 28T and 40T tokens, we decay the learning rate to $2.6 \times 10^{-5}$ following a cosine schedule.
We keep the learning rate at this value from 40T to 45T tokens.
We train the model from scratch with sparse attention at a sequence length of 64K and extend the sequence length to 1M at 34T tokens.
For auxiliary-loss-free load balancing, we set the bias update speed to 0.001 for both image and text tokens, while retaining a small sequence-level balance loss with a loss weight of 0.0001 to avoid extreme imbalance within single sequences.
Similar to DeepSeek-V4, we employ sample-level attention masking during pre-training.

\paragraph{Vision Encoder Training.} 
Our DeepSeek-ViT undergoes a separate training stage before being integrated with the language backbone. The training pipeline consists of two stages: \textit{contrastive pre-training} and \textit{autoregressive fine-tuning}. During contrastive pretraining, we optimize the model using the sigmoid contrastive loss introduced by SigLIP~\citep{zhai2023sigmoid} on approximately 47B image-text pairs sourced from alt-text data. To efficiently learn visual representations from such massive datasets, we restrict the maximum input resolution to $224 \times 224$ pixels by downscaling larger images while preserving their aspect ratios. Although using higher resolutions in this phase yields notable gains, empirical results show that these benefits contribute little to the final model. Because the subsequent autoregressive stage specifically handles high-resolution extrapolation, scaling up resolutions during contrastive pretraining significantly increases computational overhead without much overall improvement.
In the autoregressive fine-tuning stage, we connect the vision encoder to a 4B MoE LLM and train on 236B tokens across datasets including image captions, alt text, charts, and OCR, using a next-token prediction objective. This stage aims to enhance the encoder's ability to model fine-grained visual features. We therefore constrain the input resolution between $544 \times 544$ and $1344 \times 1344$ pixels by proportionally scaling out-of-bound images. After this stage, we discard the LLM and retain only the optimized vision encoder for the subsequent pre-training pipeline, where the same input-resolution policy is maintained.

\subsection{Evaluations}

\subsubsection{Evaluation Benchmarks}
We compare DeepSeek-V4.1-Flash-Base with its predecessor models DeepSeek-V4-Flash-Base and DeepSeek-V4-Pro-Base.  We report benchmarks spanning five key dimensions: world knowledge, language understanding and reasoning, coding and mathematics, long context, and multimodal abilities.

\textbf{World knowledge} benchmarks include 
AGIEval~\citep{agieval}, MMLU-Pro~\citep{mmlu_pro}, C-Eval~\citep{ceval}, 
MultiLoKo~\citep{multiloko},
SimpleQA-Verified~\citep{haas2025simpleqa} and SuperGPQA~\citep{du2025supergpqa}，

\textbf{Language understanding and reasoning} benchmarks include 
BigBench Hard (BBH)~\citep{bbh}, BigBench Extra Hard (BBEH)~\citep{kazemi2025big}, DROP~\citep{drop} and HellaSwag~\citep{hellaswag},

\textbf{Coding and mathematical} benchmarks include
BigCodeBench~\citep{big_code_bench}, HumanEval~\citep{codex}, GSM8K~\citep{gsm8k}, MATH~\citep{hendrycks2021measuring} and MGSM~\citep{mgsm}, 

\textbf{Long context} benchmark includes LongBench-V2~\citep{bai2025longbench}.

\textbf{Multimodal} benchmarks include MMMU-Pro~\citep{yue2025mmmu}, DocVQA~\citep{mathew2021docvqa}, CVBench~\citep{tong2024cambrian1} and RefCOCO/RefCOCO+/RefCOCO-g~\citep{kazemzadeh2014referitgame,nagaraja2016modeling,mao2016generation,yu2016modeling}.

\begin{table}[!h]
    \centering
    \footnotesize
    \caption{
        Comparison among DeepSeek-V4-Flash-Base, DeepSeek-V4-Pro-Base, and DeepSeek-V4.1-Flash-Base.
        All models are evaluated in our internal framework and share the same evaluation setting.
        Scores with a gap not exceeding 0.3 are considered to be at the same level.
        The highest score in each row is in \textbf{bold font}, and the second is \underline{underlined}. 
    }
    \setlength{\tabcolsep}{4pt}
    \begin{tabular}{@{}c l c | c c c@{}}
    \toprule
    & \multirow{2}{*}{\centering \textbf{Benchmark {\tiny (Metric)}}}
    & \multirow{2}{*}{\textbf{\# Shots}}
    & \textbf{DeepSeek-V4-Flash}
    & \textbf{DeepSeek-V4-Pro}
    & \textbf{DeepSeek-V4.1-Flash} \\
    & & & \textbf{Base} & \textbf{Base} & \textbf{Base} \\
    \midrule
    & Architecture & - & MoE & MoE & MoE\\
    & \# Activated Params & - & 13B & 49B & 8B/16B \\
    & \# Backbone Params & - & 284B & 1.6T & 552B \\
    \midrule
    \multirow{6}{*}{World Knowl.}
    & AGIEval {\tiny (EM)} & 3-5-shot & \underline{83.9} & \textbf{84.4} & 83.4 \\
    & MMLU-Pro {\tiny (EM)} & 5-shot & 68.3 & \underline{73.5} & \textbf{74.1} \\
    & C-Eval {\tiny (EM)} & 5-shot & \underline{92.1} & \textbf{93.1} & \underline{92.1} \\
    & MultiLoKo {\tiny (LLM-Judge)} & 5-shot & 42.6 & \textbf{50.9} & \underline{45.5}\\
    & Simple-QA verified {\tiny (EM)} & 25-shot & 30.1 & \textbf{55.2} & \underline{42.3} \\
    & SuperGPQA {\tiny (EM)} & 5-shot & 46.5 & \textbf{53.9} & \underline{53.1} \\
    \midrule
    \multirow{4}{*}{Lang. \& Reas.}
    & BBH {\tiny (EM)} & 3-shot & \underline{86.9} & \textbf{87.5} & 86.1 \\
    & BBEH {\tiny (EM)} & 1-shot & 25.4 & \textbf{29.8} & \underline{27.2} \\
    & DROP {\tiny (F1)} & 1-shot & \textbf{88.6} & \textbf{88.7} & 87.9 \\
    & HellaSwag {\tiny (EM)} & 0-shot & 85.7 & \textbf{88.0} & \underline{87.2} \\
    \midrule
    \multirow{5}{*}{Code \& Math}
    & BigCodeBench {\tiny (Pass@1)} & 3-shot & 56.8 & \underline{59.2} & \textbf{60.6} \\
    & HumanEval {\tiny (Pass@1)} & 0-shot & 69.5 & \underline{76.8} & \textbf{79.4} \\
    & GSM8K {\tiny (EM)} & 8-shot & 90.8 & \underline{92.6} & \textbf{93.0} \\
    & MATH {\tiny (EM)} & 4-shot & 57.4 & \textbf{64.5} & \underline{61.1} \\
    & MGSM {\tiny (EM)} & 8-shot & \textbf{85.7} & \underline{84.4} & 80.2 \\
    \midrule
    \multirow{1}{*}{Long Context}
    & LongBench-V2 {\tiny (EM)} & 1-shot & 44.7 & \textbf{51.5} & \underline{45.2}\\
    \midrule
    \multirow{4}{*}{Multimodal}
    & MMMU-Pro {\tiny (EM)}  & 4-shot & - & - & 56.5 \\
    & CVBench {\tiny (EM)}  & 4-shot & - & - & 77.9 \\
    & DocVQA {\tiny (LLM-Judge)} & 4-shot & - & - & 95.6 \\
    & RefCOCO-avg {\tiny (Acc@0.5)} & 0-shot & - & - &86.0 \\
    \bottomrule
    \end{tabular}
    \label{tab:base}
\end{table}

\subsubsection{Evaluation Results}

In Table~\ref{tab:base}, we provide a detailed comparison of the base models for DeepSeek-V4-Flash, DeepSeek-V4-Pro and DeepSeek-V4.1-Flash, all evaluated under our internal evaluation framework using strictly controlled and reproducible settings. 
Compared with DeepSeek-V4-Flash-Base and DeepSeek-V4-Pro-Base, our latest base model reveals a compelling efficiency gain. DeepSeek-V4.1-Flash activates a substantially smaller number of parameters than DeepSeek-V4-Pro-Base and occupies a heavily-reduced KV cache, yet its performance is fully on par with its predecessors. 
These results also reflect the substantial improvements we made to our pre-training data curation pipeline.
In this version, we introduce native multimodal training and validate its effectiveness through corresponding multimodal evaluations. Trained on a more diverse and multimodal corpus, DeepSeek-V4.1-Flash achieves world knowledge and comprehension capabilities comparable to those of DeepSeek-V4-Pro.
In reasoning and coding benchmarks, DeepSeek-V4.1-Flash shows consistent progress, reaching performance close to or better than DeepSeek-V4-Pro across multiple benchmarks.

\begin{figure}[t]
    \centering
    \includegraphics[width=0.64\textwidth]{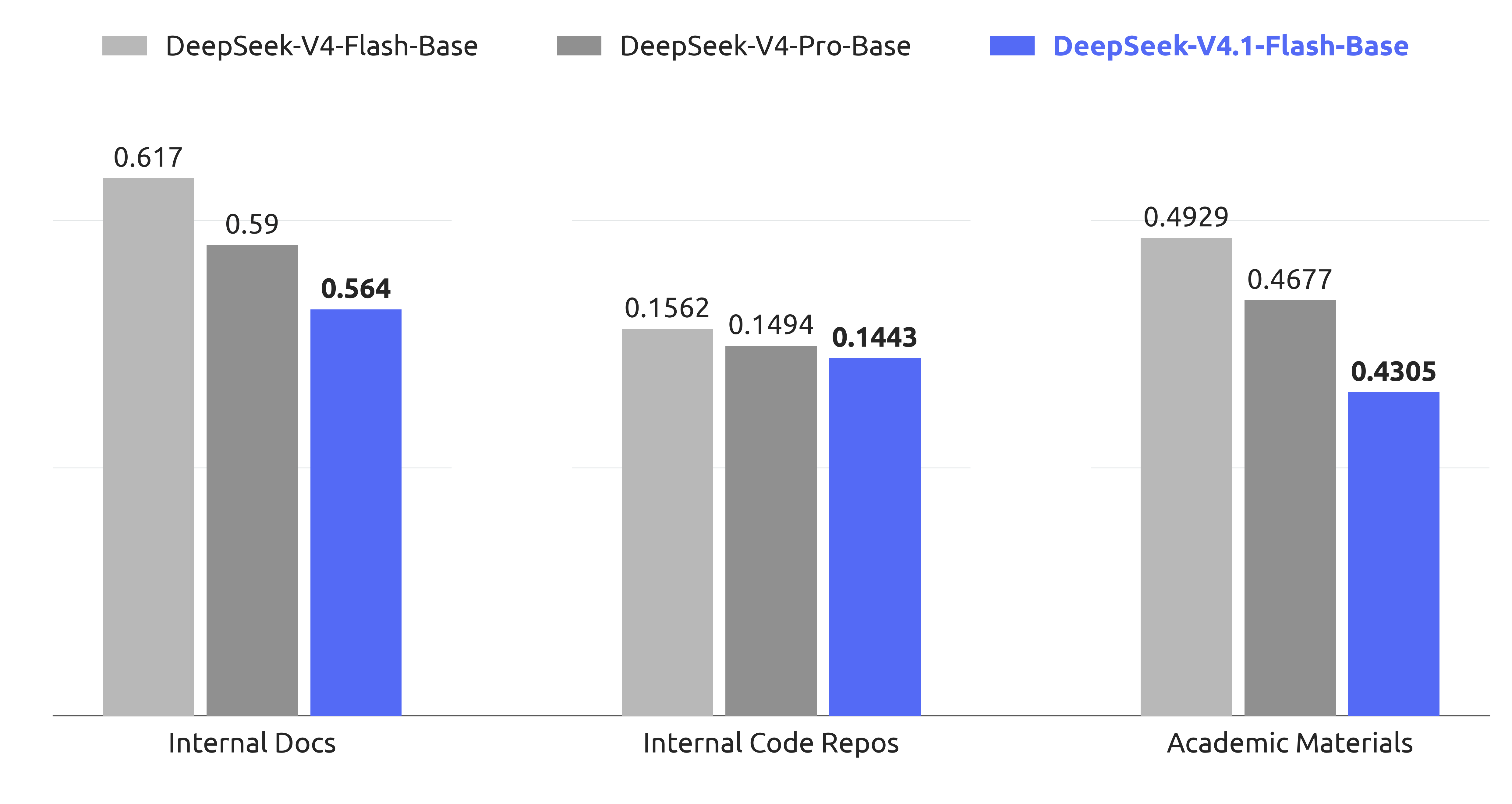}
    \caption{Bits-per-bytes (BPB) comparison of DeepSeek-V4-Flash-Base, DeepSeek-V4-Pro-Base and DeepSeek-V4.1-Flash-Base on our held-out evaluation sets. DeepSeek-V4.1-Flash-Base achieves lowest BPB on all tasks and demonstrates greater potential to serve as a strong base-model. }
    \label{fig:pretrain-eval-bpb}
\end{figure}

To further assess the model's capabilities in real-world R\&D scenarios, we additionally perform perplexity tests on dedicated internal corpora. Since perplexity tests are impossible through model APIs, we mainly focus on our own pretrained base models. For corpus selection, a separate evaluation set is collected based on our daily development, including internal documentation, proprietary code repositories and academic materials, which targets reasoning, attribution and problem-solving on complex scientific problems and frontier research. The results are shown in Figure~\ref{fig:pretrain-eval-bpb}, where we report the bits-per-byte(BPB) of different models, with lower values indicating better performance.

\section{Post-Training}
\label{sec:post-training}

\subsection{Post-Training Pipeline}
In this release, we refrain from introducing novel post-training algorithms. The overall recipe follows the standard paradigm of supervised fine-tuning (SFT) followed by reinforcement learning (RL) and on-policy distillation (OPD;~\citealp{minillm,lu2025onpolicydistillation}), without algorithmic modifications beyond well-established practices. Instead, our efforts are concentrated almost entirely on \emph{what} the model is trained on rather than \emph{how} it is optimized: we invest in large-scale, automated pipelines for data synthesis and environment construction. Concretely, the pipeline (i) synthesizes diverse, verifiable training tasks together with their reference solutions and reward signals, (ii) procedurally constructs and scales interactive agent environments in which trajectories can be collected and evaluated at low cost, and (iii) applies rigorous filtering, deduplication, and difficulty calibration to ensure data quality and curriculum balance. We find that, under a fixed and unremarkable optimization procedure, systematic improvements in the scale, diversity, and verifiability of synthesized data and environments account for essentially all of the observed gains. This observation echoes a broader lesson: at the current stage, the marginal return of engineering the data and environment pipeline substantially exceeds that of algorithmic novelty in post-training.

\subsubsection{Large-Scale Agent Task Synthesis}
Tasks serve as the fundamental fuel for agent learning. However, constructing high-quality training tasks has traditionally required substantial manual effort. We observe that the model is already beginning to exhibit the ability to construct its own training tasks, though this capability remains far from perfect. Recognizing this potential, we have invested considerable effort in strengthening the model’s task-construction and quality-verification abilities.

We formalize each task as a triplet (problem, environment, verification system) and evaluate its quality along two dimensions: difficulty—ensuring the task is non-trivial—and correctness—guaranteeing that no critical flaws exist among the three components. Using difficulty and correctness as reward signals, we iteratively train the model to construct better tasks. We also monitor RL tasks across their full lifecycle. Whenever a task is used in a new RL run, the resulting trajectories provide fresh evidence for quality re-auditing.  Under this general framework, we have built dedicated training environment production pipelines for two core scenarios: general agents and coding agents.

\textbf{General Agent.} For general agents, we encourage internal employees and external partners to incorporate our latest model into their routine workflows and, on a voluntary basis, return interaction data and feedback. Based on the interfaces observed in the returned data, we construct a large set of mocked tools that reproduce the interfaces and behaviors of real-world tools and systems, including their input formats, output structures, API schemas, and behavioral constraints, covering both commonly used SaaS and enterprise applications as well as more specialized business back-end systems. In parallel, we collect negative feedback and model failure cases submitted by internal employees at scale, and incorporate them into the pipeline to generate both single-turn and multi-turn agent environments grounded in real workflows. By reconstructing the relevant tool context, user interaction patterns, and failure conditions, the pipeline enables systematic replay of failures and targeted reinforcement learning against observed model weaknesses.

\textbf{Coding Agent.} Coding agent training environments are built from two sources: 1) coding-agent sessions from internal employees and external partners, filtered to retain highly complex tasks or tasks on which model performance is poor, then deduplicated by trajectory; and 2) public GitHub repositories that meet a star-count threshold. Environment construction is carried out collaboratively by multiple specialized agents. First, an agent determines whether the project can be built and fully run inside a container and whether it can be automatically verified; if so, it selects a specific turn or commit as the task starting point, designs several sufficiently complex implementation directions, and produces concrete evaluation points, including both fail-to-pass and pass-to-pass points, along with a construction report, fetching external resources from the web as needed. Next, a separate agent sets up dependencies, the initial working directory, test code, and task descriptions in an isolated container, performs self-testing, removes any traces that could leak the task solution, and packages the environment as a new image layer. Then, multiple distinct agents attempt the task, and an independent quality-inspection agent reviews the environment together with the solving agents' trajectories, checking for environment issues, factual errors, mismatches between evaluation points and task descriptions, and hackability risks. If the inspection does not pass, a repair agent fixes all identified errors, adjusts evaluation points that are too easy or too difficult, and the task re-enters verification.

Through these pipelines, we can automatically and batch-produce RL training data that is correct, discriminative, and controllable in length and difficulty. Whether through the faithful reconstruction of real workflows in general agent environments or the precise construction of coding tasks in coding agent environments, both ultimately feed into a unified training system, driving iterative improvement of model capabilities under continuous quality monitoring.

\subsubsection{RL in Synthesized Tasks }

\begin{figure}[t]
    \centering
    \includegraphics[width=0.9\textwidth]{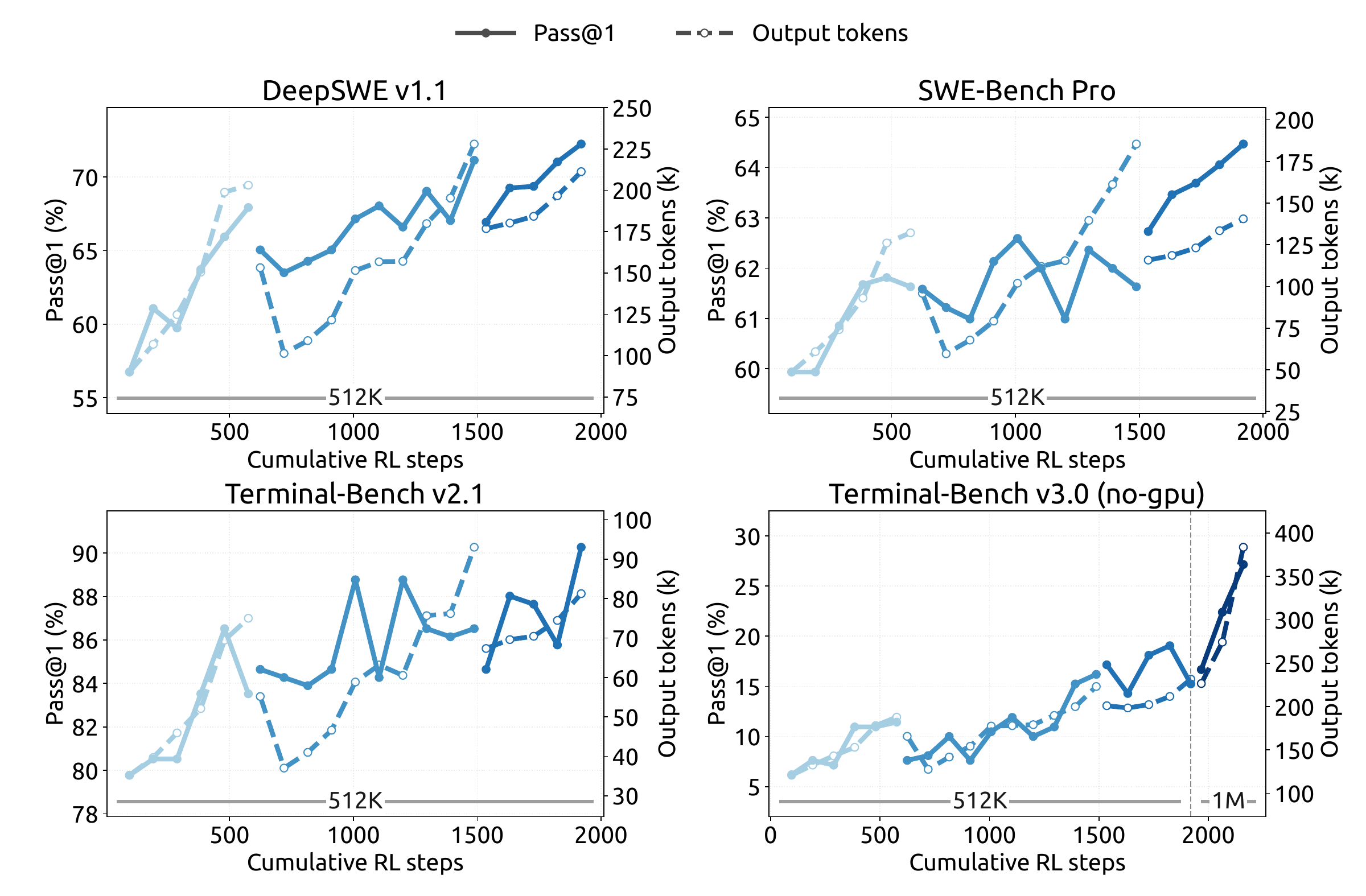}
    \label{fig:rl_scaling_cswe}
    \caption{Performance improves on various code agent benchmarks as RL training scales in the Minimal mode of DeepSeek Harness. Further extending maximal context length to 1M tokens continues to improve performance on extremely long-horizon tasks, e.g., Terminal-Bench v3.0.}
    \label{fig:rl_scaling}
\end{figure}

\begin{figure}[t]
    \centering
    \includegraphics[width=0.9\textwidth]{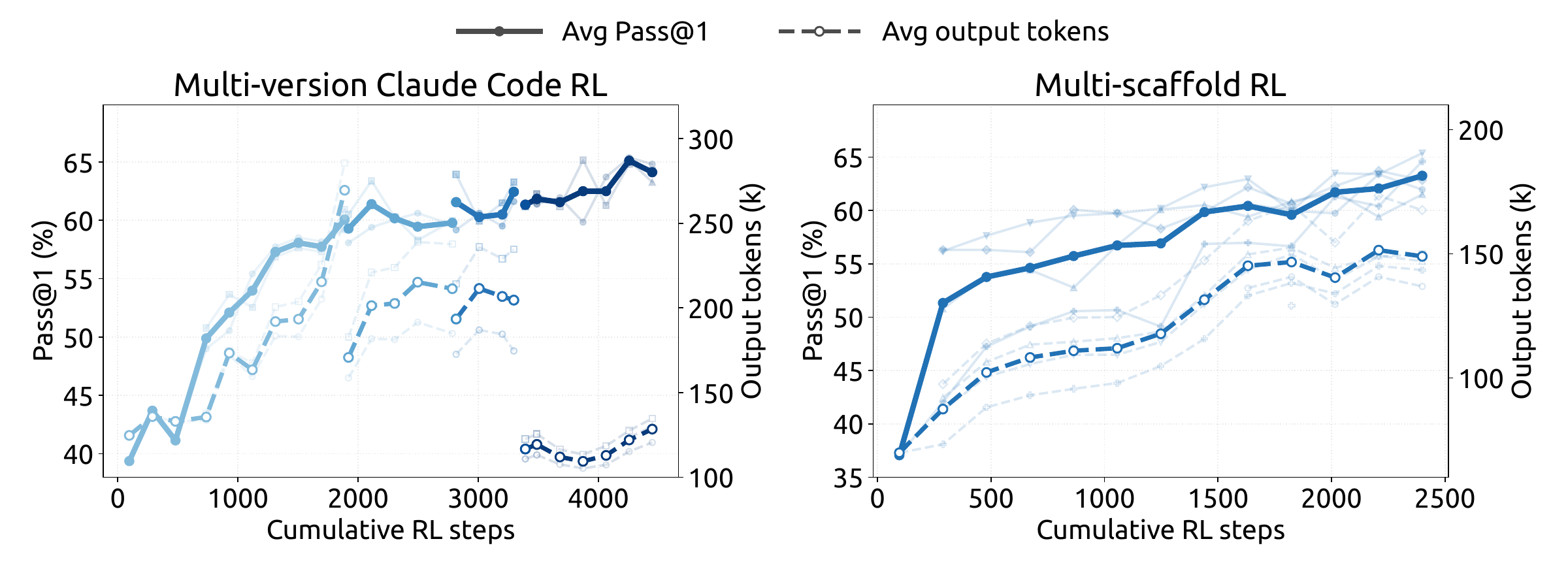}
    \label{fig:rl_scaling_cc_multi}
    \caption{
    Performance improves with cumulative RL steps when jointly training across multiple versions of Claude Code (left) and across heterogeneous scaffolds, including OpenCode, Pi, and DeepSeek Harness in Standard and PTC modes (right). Performance is evaluated on DeepSWE v1.1. Lighter curves show the evaluation of individual scaffold versions or scaffolds.
    }
    \label{fig:rl_scaling_multi}
\end{figure}

We use large-scale asynchronous RL in synthesized tasks to improve model performance and shape its behavior in complicated scenarios. We scale RL runs in two dimensions, i.e., training compute and the number of scaffolds. 
As shown in Figure~\ref{fig:rl_scaling} and Figure~\ref{fig:rl_scaling_multi}, performance continues to improve as we increase compute with cumulative RL steps, whether scaling within a single scaffold, jointly across variants of the same scaffold, or across heterogeneous scaffolds.

For RL training across diverse scaffolds, we decouple agent rollout execution into an agent sandbox and a worker container. 
The sandbox runs the scaffold and its tools, while the worker provides a scaffold-agnostic control layer that orchestrates the rollout, normalizes heterogeneous interactions into a common trajectory schema, and communicates with the trainer. 
Both run on DSec (Section~\ref{sec:dsec}), outside the preemptible GPU training pool, separating long-lived rollouts from fine-grained training scheduling. 
During trainer preemption, rollout execution can be suspended and offloaded while preserving its full state for later resumption and releasing CPU and GPU resources.
This design enables stable and efficient RL across heterogeneous scaffolds without modifying the underlying algorithms. 

To extend effective RL compute beyond a single training run, we use model merging to reinitialize successive RL runs. 
Specifically, we merge checkpoints from runs across different scaffolds or configurations, combining improvements acquired along different optimization paths. 
In Figure~\ref{fig:rl_scaling} and Figure~\ref{fig:rl_scaling_multi}, disconnected curve segments reflect successive RL runs after model reinitialization. 
This yields further gains in both task performance and token efficiency, providing a simple and practical way to aggregate parallel RL compute and continue scaling across successive runs.

\subsubsection{Running Agents at Massive Scale: DSec}
\label{sec:dsec}

As we transitioned from DeepSeek-V3 to V4, the rapidly growing number and diversity of agentic training environments motivated us to build \textit{DeepSeek Elastic Compute (DSec)}, a production-grade sandbox platform for large-scale agentic training and evaluation. Its initial design addressed heterogeneous execution environments, scalable image distribution, multiple isolation backends, high-density resource management, command trajectory logging, and preemption-safe resumption.

V4.1 training further increased demand to millions of concurrent sandbox instances spanning diverse harnesses, platforms, code repositories, software dependencies, and task-specific services. At this scale, the primary bottlenecks shifted toward datacenter scalability, workload isolation, per-node compute density, and the containment of misbehavior by increasingly capable agents. We briefly describe our key design aspects below.

\textbf{Horizontally Scaling Compute at Scale.} DSec scales through two complementary mechanisms: sharding and scheduling with relaxed consistency. To accommodate a large number of machines, we partition compute nodes into multiple shards (so-called scale units). Such sharding also allows us to reduce blast radius by isolating workloads from different experiments, preventing a single memory-intensive task from exhausting resources shared by unrelated work.

Instead of using off-the-shelf orchestrators like Kubernetes, DSec employs a custom placement engine to schedule the large number of sandboxes, by trading strong global consistency for scalability. This design is based on a key observation: agentic sandbox placements only demand eventual consistency as long as each compute node enforces local safety constraints. To do so, the placement engine deploys in multiple independent replicas, without synchronized coordination. Each replica predicts resource availability from recent measurements and makes good-enough placement decisions. To compensate for this loss in consistency, each node is responsible for validating the final placement decisions, enforcing a hard admission constraint that rejects new placements if it exceeds a local warning threshold. This design as a whole allows DSec to scale to millions of containers without bottlenecking on central coordination.

\textbf{Running Sandboxes at High Density.} At the node level, we use hardware-supported sub-NUMA partitioning and bind each worker VM to an individual NUMA domain. Containers run within these worker VMs, with their CPU and memory allocations confined to the VM's local NUMA resources. This setup balances aggressive memory overcommitment against Linux kernel lock contention, while localizing memory pressure and runtime failures. Under comparable workload configurations, it increases the supported density from roughly 1,000 to more than 2,500 concurrent live containers per physical node before measurable end-to-end degradation appears.

Such high-density deployment can nevertheless distort time-sensitive evaluations through interference from background workloads. DSec therefore introduces a latency-sensitive (LS) execution class. We apply \texttt{SCHED\_IDLE} to non-LS tasks to minimize their scheduling priority, and use core scheduling to ensure only tasks of the same priority class execute simultaneously on sibling hyperthreads to eliminate interference.

\textbf{Mitigation of Misbehaving Agents.} During RL training, we frequently observe agents attempting to perform reward hacking or unintentionally crashing the environment. 
In certain attempts, our agents exploited recently disclosed vulnerabilities, including permission issues from the XFS driver, illegal memory access in AppArmor, leaking answers from package mirror services, and so on. Agents have also been notorious for deleting critical binaries, breaking system files, or even removing the filesystem.
We use per-sandbox AppArmor profiles and fine-grained eBPF-based network policies to prevent such attempts. If an agent crashes its environment, we treat the crash as a failed trajectory and report a "repercussion" signal to the RL framework.

\subsubsection{Controllable Reasoning Effort in RL}

Alongside advances in model architecture and hardware, the number of
output tokens is another key
determinant of serving cost and, consequently, of the cost--quality
trade-off in real-world applications. We therefore introduce a scalar
effort level $b$ as an explicit conditioning signal during
reinforcement-learning. This mechanism is applied to both
single-turn reasoning and multi-turn agentic tasks. Specifically, we
prepend the following instruction to the system prompt:
\begin{center}
    \texttt{Reasoning Effort: \{effort\} (range 1--100; higher values request more thorough reasoning)}
\end{center}
Here, $b\in\{1,\ldots,100\}$ denotes the requested effort level.

For each training prompt $x$, we sample $M_b$ responses at each effort
level $b\in\mathcal{B}$:
\begin{equation}
    z_{b,j}
    \sim
    \pi_{\theta}(\cdot\mid x,b),
    \qquad
    b\in\mathcal{B},
    \quad
    j=1,\ldots,M_b .
    \label{eq:effort-conditioned-rollout}
\end{equation}
Here, $j$ indexes the responses sampled at effort level $b$. Responses
sharing the same $(x,b)$ form a subgroup, within which rewards are
mean-centered to compute group-relative advantages. Thus, responses
from different effort levels are not directly compared. Instead,
effort-dependent behavior is induced within each subgroup by making
the length component of the reward depend on $b$. Specifically, we add
the length-penalty term $r_{b,j}^{\mathrm{len}}$ to the reward of response $z_{b,j}$:
\begin{equation}
    r_{b,j}^{\mathrm{len}}
    =
    -\min\left\{
        C_{\max},
        \;
        k(b)\frac{\ell_{b,j}}{L_{\mathrm{norm}}}
    \right\},
    \label{eq:length-penalty}
\end{equation}
where $\ell_{b,j}$ is the number of reasoning tokens,
$L_{\mathrm{norm}}$ is a reference length, and $C_{\max}$ caps the
maximum deduction applied to a trajectory. The token-penalty coefficient decreases exponentially with the
requested effort:
\begin{equation}
    k(b)
    =
    k_0
    \exp\left(
        -\frac{b-b_{\min}}{\tau}
    \right),
    \qquad
    \tau
    =
    \lambda\overline{\Delta b},
    \label{eq:token-penalty}
\end{equation}
where $k_0$ is the basic penalty coefficient at different effort
levels, $b_{min}$ is the minimum value of $\mathcal{B}$, $\overline{\Delta b}$ is the average spacing between training effort
levels, and $\lambda$ controls the rate of penalty decay. Increasing
$b$ by $\tau$ multiplies the penalty coefficient by $e^{-1}$. The
parameter $k_0$ controls the overall pressure toward shorter reasoning,
whereas a smaller $\tau$ causes the penalty to decay more rapidly and
tends to produce greater behavioral separation between effort levels.
Appendix~\ref{app:effort-penalty} provides a marginal-utility motivation
for the exponential parameterization of $k(b)$.

At deployment time, the scalar $b$ provides a flexible control
interface over the model's reasoning strength. By varying $b$, a single model checkpoint can move between
different operating regimes along the learned cost--quality frontier,
adapting to different latency, token-budget, and solution-quality
requirements. Although training uses only a finite set of
effort levels, intermediate values can be used at deployment to elicit
interpolated reasoning behaviors, providing a fine-grained and efficient
mechanism for test-time resource allocation.

In our production deployment launched in September 2026, the
public API exposes three preset reasoning-effort tiers---\texttt{max},
\texttt{high}, and \texttt{low}---which map directly onto this scalar
interface. As summarized in Table~\ref{tab:api-effort-tiers}, the three
tiers correspond to effort values of $b=100$, $b=75$, and $b=50$,
respectively, so that API users select an operating point on the learned
cost--quality frontier without any change to the model weights or
decoding configuration.
\begin{table}[htbp]
    \centering
    \caption{Mapping between the public API reasoning-effort tiers and the underlying scalar effort values $b$.}
    \label{tab:api-effort-tiers}
    \begin{tabular}{lc}
        \toprule
        API tier & Effort value $b$ \\
        \midrule
        \texttt{max}  & 100 \\
        \texttt{high} & 75  \\
        \texttt{low}  & 50  \\
        \bottomrule
    \end{tabular}
\end{table}

\subsection{Asynchronous Post-training Infrastructure}

The long-tail problem during the rollout phase of RL for LLMs has consistently been a major bottleneck for training efficiency. To address this, we extend our post-training infrastructure to allow asynchronous generation of samples~\citep{zeng2026glm,K2.5}, which significantly mitigates the long-tail issue in the rollout phase by maintaining a sufficiently high level of concurrency. Asynchronous training is now enabled for nearly all our RL and OPD tasks, and rollout efficiency has improved substantially.

\subsubsection{Overall Workflow}

We colocate rollout and training on the same physical devices and time-share their execution, eliminating the need to manually tune resource allocation between the two phases. Each task specifies an upper bound on the number of in-flight samples, and the system maintains this bound throughout the rollout phase.

We evaluated three dispatch granularities for maintaining the target rollout concurrency.
Our final approach is sample-level dispatch: 
once the number of newly completed samples reaches the GRPO group size assigned to the next prompt, we dispatch that prompt regardless of which groups produced those completions. 
This helps maintain a steady rollout concurrency throughout training. Before settling on this dispatch strategy, we experimented with two alternative dispatch granularities. In the first attempt, we dispatched several extra batches at the beginning and supplemented a full batch after each training iteration; however, this caused severe oscillations in training metrics, indicating that batch-level granularity was too coarse. In the second attempt, we switched to prompt-level dispatch, where a new prompt was dispatched after one GRPO group finishes, but found that it stalled easily on long-tail samples within a GRPO group, making it difficult to smoothly maintain the target rollout concurrency.

Once enough training samples have accumulated, training preempts ongoing rollouts. During training, we use concatenated routing-replay: for samples that span multiple checkpoints, we concatenate the expert routing produced at each rollout segment rather than discarding and recomputing the routing information with new checkpoints.

\subsubsection{Mitigating Length Bias and Off-Policy Effects}

Asynchronous generation, while effectively improving rollout efficiency, introduces two side effects that can degrade training quality. First, it creates a length-distribution bias, especially in the early training stage, because shorter sequences tend to complete first and thus dominate the initial training batches. Second, it inevitably produces off-policy samples, i.e., samples whose tokens are partially or entirely generated by earlier checkpoints. These two issues require different handling strategies.

To address the length bias, we employ two mechanisms. First, the dispatcher can limit concurrency on a per-dataset basis, which helps regulate the proportion of each dataset in the steady-state training batch, indirectly mitigating the length skew by controlling the sources of incoming samples. Second, we support discarding early-returned short samples to smooth the transition into the steady-state length distribution, and prevent the model from overfitting to overly short sequences.

For the off-policy issue, we implement two additional mechanisms. First, by tuning the logic that controls sample dispatching and the waiting condition for training samples, we can bound the maximum off-policy ratio, ensuring that the training data does not deviate excessively from the current model. Second, during training, we add a loss masking scheme that eliminates the contribution of tokens with excessive staleness, thereby mitigating the adverse impact of stale samples on gradient updates.

\subsubsection{Performance Optimization}

In our asynchronous RL framework, the rollout phase is periodically interrupted to switch to updated policy checkpoints. We aim to make this process seamless: interruptions should be near-instantaneous, and interrupted rollouts should resume as if never stopped.

To stop rollouts promptly, we support token-level interruption: generation can be halted at any token boundary. Once sufficient training data has been collected, all in-flight samples stop almost immediately, allowing the system to enter the training phase without delay.

To preserve rollout progress across checkpoint switches, rollout states such as KV cache and expert routing are persisted at token granularity during generation. Upon resumption with a new checkpoint, the persisted states are directly reused, eliminating the cost of re-prefilling and allowing interrupted samples to continue exactly where they left off. Since this requires retaining the states of all in-flight samples, we perform sample-grained garbage collection, releasing each sample's states as soon as it completes.

Beyond checkpoint switching, the same fast-interruption and seamless-resumption machinery allows the training jobs to respond promptly to cluster scheduling preemption signals without losing progress, thereby improving overall cluster utilization.

\subsubsection{Large-Scale On-Policy Distillation}

As the last stage of post-training, the final full-vocabulary OPD task is trained on datasets from all domains using over 40 teacher models. It also adopts asynchronous generation to improve rollout efficiency. Due to differences in training procedures across domains, the best teacher for each domain may come from a different stage of model development. Moreover, the teacher models may differ architecturally from one another and from the student. Our post-training infrastructure readily accommodates this setting, supporting full-vocabulary OPD with an effectively unbounded number of architecturally heterogeneous teachers, and efficient switching among them at negligible cost~\citep{dsv4}.

The OPD stage also requires dynamic reconfiguration during training. 
We continuously track model capabilities and may adjust the training recipe accordingly, including the dataset mixture, per-dataset concurrency limits, and active teachers.
Such changes are straightforward in synchronous training where rollout batches provide explicit configuration boundaries. In the asynchronous setting, however, samples generated under different configurations may coexist in flight. Our infrastructure supports consistent transitions between configurations without disrupting rollout or training.

\subsection{Evaluation}

\subsubsection{Evaluation Setup}

Our post-training evaluation focuses primarily on reasoning and agentic capabilities, while knowledge-intensive performance is largely determined by pretraining and reported in Table~\ref{tab:base}.
For reasoning, we evaluate on GPQA Diamond~\citep{gpqa}, Humanity's Last Exam~\citep{hle}, Codeforces (internal benchmark), and MathArena Apex~\citep{matharenaapex}, using temperature and top-$p$ of 1.0.
For agentic capabilities, we evaluate across four categories:
\begin{itemize}
    \item \textbf{Code agent}: Terminal-Bench 2.1~\citep{merrill2026terminal}, Terminal-Bench 3.0~\citep{marten2026terminalbench3}, Terminal-Bench 4.0~\citep{Marten_Terminal-Bench_2026}, DeepSWE v1.1~\citep{deepswe}, ProgramBench~\citep{yang2026programbenchlanguagemodelsrebuild}, NL2Repo-Bench~\citep{ding2025nl2repo}.
    \item \textbf{Cyber security}: SEC-Bench Pro version 260505~\citep{lee2026sec}, CyberGym~\citep{wang2026cybergym}, and ExploitGym~\citep{wang2026exploitgym}.
    \item \textbf{General agent}: the public evaluation set of AutomationBench v1.0.6~\citep{shepard2026automationbench}, Agents' Last Exam~\citep{sun2026agents} (ALE-CLI).
    \item \textbf{Visual agent}: Chartography~\citep{garre2026chartography}, BabyVision~\citep{chen2026babyvision}, main set of ZeroBench~\citep{roberts2025zerobench}.
\end{itemize}
For code agents, we evaluate DeepSeek-V4.1-Flash using the Minimal mode of DeepSeek Harness with a 1M-token context window, temperature set to 1.0, and top-p set to 0.95. 
To align with official setup requirements, we employ the mini-SWE harness for DeepSWE v1.1. For SEC-Bench Pro, we utilize the Claude Code harness specifically for its session compact design.
For visual agent tasks, we evaluate using the Claude Code harness with a 512k-token context window, temperature set to 1.0, and top-p set to 0.95.
Agents' Last Exam and AutomationBench are evaluated with their official scaffolds. Model performance with other coding scaffolds is reported in Table \ref{tab:agent-scaffolds}.

To mitigate reward hacking in coding agent evaluations, we restrict internet access and strip Git histories from the environment. Additionally, we automatically purge transient build and package caches across diverse environments, including Go module caches (go/mod), node modules dependency artifacts, compiled .jar files, and Python pycache directories. Despite these precautions, we still observe instances of exploit-seeking behavior during testing—such as decompiling core Ubuntu Linux packages to uncover vulnerabilities in CyberGym. As models grow increasingly capable, standard evaluation infrastructure (e.g., Docker containers and validation scripts) becomes more susceptible to model gaming. We urge the broader research community to prioritize detecting and mitigating these behaviors when designing next-generation benchmarks.

\subsubsection{Evaluation Results}
\begin{table}[ht]
    \centering
    \footnotesize
    \setlength{\tabcolsep}{1.2pt}
    \caption{
        Comparison between DeepSeek-V4.1-Flash with closed/open source models.  $\dag$ denotes text-only subset of HLE.
        The best results are highlighted in bold; the second-best results are underlined.
    }
    \label{tab:large_eval}
    \resizebox{\textwidth}{!}{
    \begin{tabular}{@{}c l | c c  c c c c | c @{}}
    \toprule

    & \multirow{2}{*}{\centering \textbf{Benchmark {\tiny (Metric)}}} & \textbf{Opus-5} & \textbf{GPT-5.6 Sol} & \textbf{K3} & \textbf{GLM-5.3} & \textbf{DS-V4-Pro} & \textbf{DS-V4-Flash} & \textbf{DS-V4.1-Flash} \\
    & & \textbf{Max} & \textbf{Max} & \textbf{Max} & \textbf{Max} & \textbf{Max} & \textbf{Max} & \textbf{Max} \\

    \midrule

    \multirow{4}{*}{\rotatebox{90}{Reasoning}}
    & GPQA Diamond {\tiny (Pass@1)}       &\underline{93.4} &\textbf{94.1} &92.9 &88.1 &92.4 &89.9 &90.9 \\
    & HLE {\tiny (Pass@1)}                &\textbf{56.3} &\underline{44.5} &43.5 &42.0$^\dag$ &42.7$^\dag$ &37.8$^\dag$ &36.8 (39.1$^\dag$) \\
    & Codeforces {\tiny (Rating)}         &- &- &- &- &\underline{3348} &3289 &\textbf{3471} \\
    & MathArena Apex {\tiny (Pass@1)}               &- &- &\textbf{65.6} &- &\underline{65.3} &58.6 &\textbf{65.6} \\
    \midrule

    \multirow{16}{*}{\rotatebox{90}{Agentic}}
    & Terminal-Bench 2.1 {\tiny (Pass@1)}  & \underline{89.1} & 88.8 & 88.3& 88.2& 87.9 & 82.7 & \textbf{90.6}\\
    & Terminal-Bench 3.0 {\tiny (Pass@1)}  & \textbf{43.3} & \underline{34.4} & 17.7& 28.3 & 11.8 & 7.6 & 30.0\\
    & Terminal-Bench 4.0 {\tiny (Pass@1)}  & \textbf{51.8} & \underline{39.9} & 12.6 & 37.9 & 12.4 & 7.0 & 31.2 \\
    & DeepSWE v1.1 {\tiny (Resolved)}  & \underline{74.0} & 73.0 & 67.5& 66.9& 62.7 & 54.4 & \textbf{74.2} \\
      & ProgramBench {\tiny (Almost@1)}  & \textbf{37.0} & \underline{23.0} & 17.5 & 19.0 & 15.5 & - & 20.3\\
    &  NL2Repo-Bench {\tiny (Score)}  & \textbf{75.3} & 56.8 & 58.0 & 58.0 & 61.5 & 54.2 & \underline{65.4} \\
    & CyberGym {\tiny (Pass@1)}  & - & \underline{84.5} &80.0 & \underline{84.5}& 83.3& 76.7 & \textbf{88.1}\\
    & SEC-Bench Pro {\tiny (Pass@1)}  & - & \textbf{74.3} &  - & - & 56.4  & 30.9   & \underline{62.8}\\
    & ExploitGym {\tiny (Pass@1)}  & \underline{22.1} & \textbf{33.7} & - & 15.0 & 5.4 &  1.8 & 15.3\\ 
  
    & HLE w/ tools {\tiny (Pass@1)}  & \underline{63.6} & - & 59.8& 62.5& 60.0 & 51.5& \textbf{63.9} \\
    & Automation-Bench {\tiny (Pass@1)} & \underline{50.3} & 45.8 &46.7 &48.8 & 43.2 & 37.7 & \textbf{54.8} \\
    & Agents' Last Exam{\tiny (Pass@1)} & \underline{28.6} & 26.7 & 27.6& 28.5& 25.7 & 25.2 & \textbf{31.8} \\

    & Chartography w/ tools {\tiny (Pass@1)}  & \textbf{84.0} & \underline{79.9} & 68.1 & -  &  - &  - & 78.9 \\
    & BabyVision w/ tools {\tiny (Pass@1)}  & \textbf{94.1} & 88.9 & 85.7 & - & - & - & \underline{89.6} \\
    & ZeroBench-main w/ tools {\tiny (Pass@5)}  & \underline{52.0} & \textbf{53.0} & 41.0 & - & - & - & 49.0 \\
    \bottomrule
    \end{tabular}}
\end{table}

As detailed in Table \ref{tab:large_eval}, DeepSeek-V4.1-Flash exhibits significant performance upgrades across both reasoning and agentic benchmarks over its predecessor, DeepSeek-V4-Flash, while matching or outperforming top-tier open-source and proprietary models.

In core reasoning tasks, DeepSeek-V4.1-Flash achieves a Codeforces rating of 3471, surpassing both DeepSeek-V4-Flash (3289) and DeepSeek-V4-Pro (3348). On MathArena Apex, it obtains a 65.6\% Pass@1 accuracy, fully matching the top-performing open-source baseline Kimi-K3 (65.6\%) and DeepSeek-V4-Pro (65.3\%). Furthermore, its GPQA Diamond score reaches 90.9\%, showing steady improvements over DeepSeek-V4-Flash (89.9\%).

The performance gains are even more pronounced across agentic tasks. Notably, on DeepSWE v1.1, DeepSeek-V4.1-Flash reaches 74.2\% pass rate, marking a substantial jump from DeepSeek-V4-Flash (54.4\%) and surpassing leading proprietary models including Opus-5 (74.0\%) and GPT-5.6 Sol (73.0\%). On Terminal-Bench 2.1, it achieves 90.6\%, outperforming Opus-5 (89.1\%) and GLM-5.3 (88.2\%). Similarly, on Automation-Bench (54.8\%) and Agents' Last Exam (31.8\%), DeepSeek-V4.1-Flash establishes leading scores over both open-source counterparts and top closed-source systems. Despite its compact nature, DeepSeek-V4.1-Flash demonstrates state-of-the-art agentic capabilities, substantially closing the gap with frontier closed-source models while establishing clear advantages among open-source alternatives.

On cyber-security tasks, DeepSeek-V4.1-Flash establishes a new state of the art among open-source models. Given the dual-use nature of these capabilities, we encourage the community to apply them responsibly, such as for defensive security research and vulnerability remediation. In the domain of visual agent tasks, DeepSeek-V4.1-Flash demonstrates robust capabilities, particularly in scenarios requiring visual reasoning and the analysis of complex professional charts. While it outperforms the leading open-source model Kimi-K3, we acknowledge that a measurable gap still remains when benchmarked against the leading closed-source alternatives.

Beyond peak performance, DeepSeek-V4.1-Flash exposes a reasoning-effort setting that allows
users to trade inference cost for accuracy in a controllable manner. As illustrated in
Figure~\ref{fig:reasoning-efforts}, both accuracy and output length increase steadily with
the effort level across reasoning and agentic benchmarks alike. Raising the effort from 25 to
100 improves the average Pass@1 on eight reasoning-intensive benchmarks from 67.1\% to 76.3\%,
on DeepSWE v1.1 from 66.0\% to 74.2\%, and on Terminal-Bench 2.1 from 82.4\% to 90.6\%, at
the cost of roughly $2.5\times$ more output tokens. Notably, the effort control learned on
single-response reasoning transfers faithfully to long-horizon agentic trajectories, where it
governs the total amount of exploration and verification across turns. The gains are
front-loaded: the 60--80 range already recovers most of the accuracy of the maximum setting
at less than half of its token budget, whereas the final step to effort 100 lengthens agent
trajectories by $1.6$--$1.8\times$ for only marginal improvements. The maximum tier is thus
best reserved for the most challenging tasks, while moderate effort levels offer a favorable
cost--performance balance for everyday agentic use.

\subsubsection{Performance across reasoning efforts}
\begin{figure}[t]
    \centering
    \includegraphics[width=1.0\linewidth]{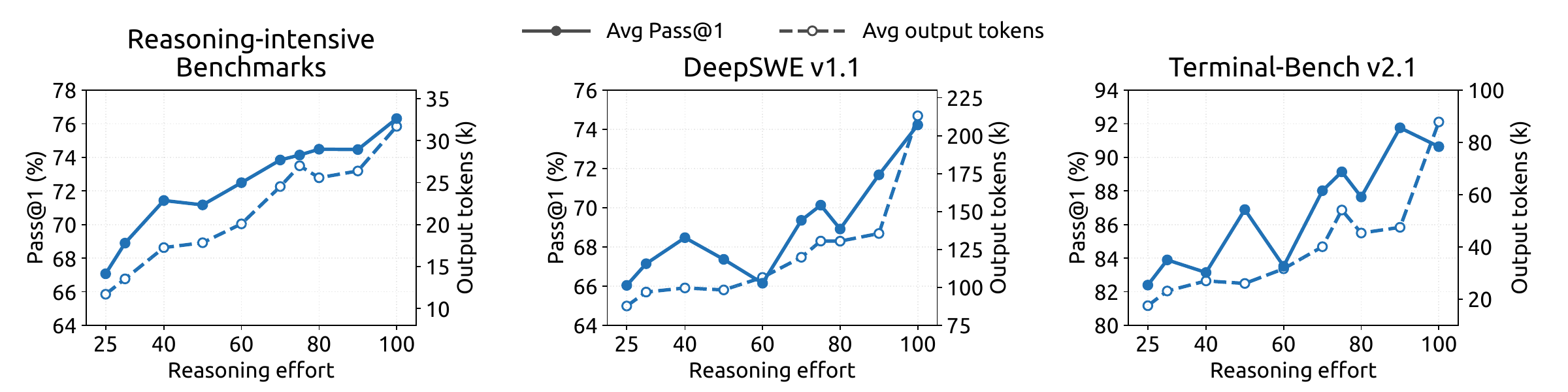}
    \caption{Performance and output length as a function of reasoning effort. Each panel plots Pass@1 (solid, left axis) and mean output
tokens per response (dashed, right axis) as the reasoning-effort value is varied from 25 to 100; The results of reasoning-intensive benchmarks are averaged over eight benchmarks (AIME 2026, Apex 2025 Shortlist, GPQA Diamond, HLE, IMO-AnswerBench, LiveCodeBench, MathArena-Apex, SimpleQA-Verified). DeepSWE v1.1 is evaluated based on mini-SWE and Terminal-Bench v2.1 is evaluated based on DeepSeek Harness (Minimal).
}
    \label{fig:reasoning-efforts}
\end{figure}

Figure~\ref{fig:reasoning-efforts} reports performance and average response length across a range of budget values. As the reasoning effort increases, the model produces progressively longer reasoning traces and accuracy improves monotonically on reasoning-intensive benchmarks and software engineering tasks, with the largest gains concentrated in the low-to-mid budget range and diminishing returns beyond. Although our RL involves only a limited number of effort levels, the use of scalar efforts achieves flexible, interpolated control of response length within a specific range. This allows practitioners to trade off quality against latency and token cost along a smooth continuum: latency-sensitive applications can operate at low effort with modest accuracy degradation, while difficult tasks can invoke high effort to recover the model's full reasoning capability. In our public API service, we expose three preset effort levels that map onto this scale: \texttt{low}, \texttt{high}, and \texttt{max} correspond to effort values of 50, 75, and 100, respectively.

\subsubsection{Performance across agent scaffolds}

In practice, a model is rarely deployed within a single fixed agent framework; different scaffolds vary in their system prompts, tool definitions, context management strategies, and interaction protocols, and a model that overfits to one particular harness may degrade substantially when placed in another. To assess the robustness of our model to such variation, our comparison covers eight configurations from six scaffold families: Claude Code~\citep{claude_code}, Codex~\citep{codex_harness}, OpenCode~\citep{opencode}, Pi~\citep{pi_agent_harness}, mini-SWE~\citep{miniswe}, and DeepSeek Harness (DSH)~\citep{deepseek-harness2026} in Minimal, Standard, and PTC modes. For each scaffold, we keep the model checkpoint, decoding configuration, and task set identical, and only the surrounding harness, including its native system prompt, tool schema, and turn-taking logic, is changed. Table~\ref{tab:agent-scaffolds} reports performance  at Max reasoning effort (100) on DeepSWE v1.1 and Terminal-Bench v2.1.

\begin{table}[ht]
  \centering
  \footnotesize
  \setlength{\tabcolsep}{3pt}
  \renewcommand{\arraystretch}{1.0}
  \caption{\centering Performance across agent scaffolds at Max reasoning effort.}
  \label{tab:agent-scaffolds}
  \begin{tabular*}{\textwidth}{@{\extracolsep{\fill}}l|ccccc|ccc@{}}
    \toprule
    \multirow{2}{*}{\textbf{Benchmark {\tiny (Metric)}}}
      & \multirow{2}{*}{\textbf{Claude Code}}
      & \multirow{2}{*}{\textbf{Codex}}
      & \multirow{2}{*}{\textbf{OpenCode}}
      & \multirow{2}{*}{\textbf{Pi}}
      & \multirow{2}{*}{\textbf{mini-SWE}}
      & \multicolumn{3}{c@{}}{\textbf{DeepSeek Harness}} \\
      & & & & & & \textbf{Minimal} & \textbf{Standard} & \textbf{PTC} \\
    \midrule
    DeepSWE v1.1 {\tiny (Resolved)}
      & 69.8 & 65.6 & 65.5 & 66.2 & 74.2 & 72.6 & 70.5 & 67.6 \\
    Terminal-Bench v2.1 {\tiny (Pass@1)}
      & 88.0 & 84.1 & 85.0 & 86.1 & 90.3 & 90.6 & 85.8 & 85.8 \\
    \bottomrule
  \end{tabular*}
  \par\vspace{3pt}
  \begin{minipage}{\textwidth}
    \scriptsize
    \textit{Note.} All scaffolds use $N=8$ samples per task on DeepSWE v1.1 and $N=3$ on Terminal-Bench v2.1.
    Runs use Linux containers, temperature 1.0, top-p 0.95, a 1M-token context window, and \texttt{max\_steps=500} model-generation rounds per agent on both benchmarks.
    Terminal-Bench v2.1 is evaluated without network access.
    We evaluate four Claude Code versions, with per-version results and their average reported in Appendix Table~\ref{tab:claude-code-versions}; this table reports v2.1.251.
    More scaffold-specific configurations are detailed in Appendix~\ref{app:multi-scaffold-eval}.
  \end{minipage}
\end{table}

The model's agentic capabilities transfer well across scaffold families with different prompts and tool interfaces, rather than depending on conventions specific to a particular harness. Its performance remains robust as the surrounding interaction protocol and tool abstractions change, indicating that its agentic behavior is not tightly coupled to a single scaffold design. This robustness is consistent with the diversity of environments, tool schemas, and interaction formats in our synthesized training data (Section~\ref{sec:post-training}), which is designed to encourage generalization across agent scaffolds.

\subsubsection{Multi-Agent}

\begin{figure}[t]
\centering
\includegraphics[width=0.9\textwidth]{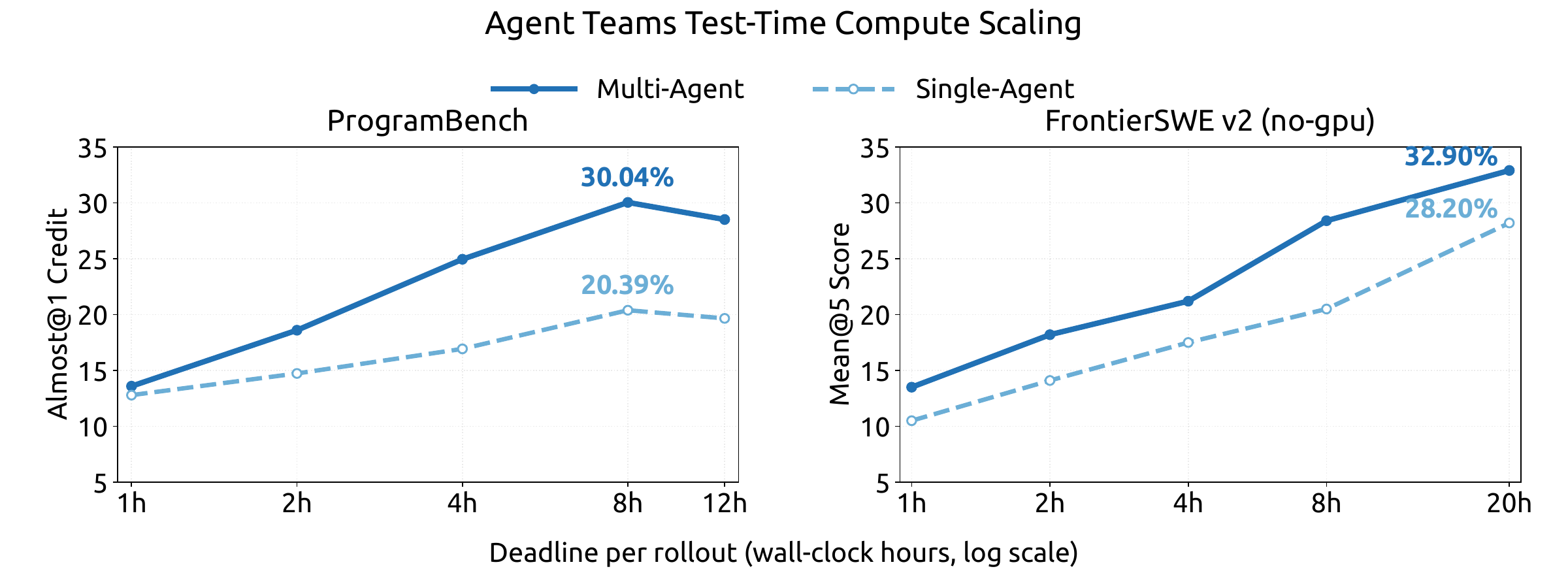}
\caption{Test-time compute scaling for single-agent and multi-agent configurations on ProgramBench (Almost@1) and FrontierSWE v2 (Mean@5) as functions of the per-rollout wall-clock deadline.}
\label{fig:multi-agent-programbench}
\end{figure}

To explore multi-agent collaboration on complex tasks, we conduct preliminary experiments with DeepSeek-V4.1-Flash using DeepSeek Harness's Agent Team mode.

\textbf{Multi-Agent Harnesses.}
We use DeepSeek Harness in Agent Team mode, where a lead agent can asynchronously create named, persistent teammates by default through \texttt{spawn\_teammate}. Each receives a delegated task and starts in either \texttt{fresh} mode without lead history or \texttt{fork} mode with a one-time snapshot of the lead's completed turns. All agents share one repository checkout, making edits immediately visible to one another.

Agents communicate through a durable peer mailbox. A message sent through \texttt{send\_message} reaches a running teammate at its next step boundary, starts a new turn for an idle teammate, or resumes an inactive teammate.
Across each teammate's task lifecycle, the lead monitors runtime status with \texttt{list\_agents} and waits for status, mailbox, or shared-task changes with \texttt{wait\_agent}. Task ownership, dependencies, and advisory write scopes are maintained on a shared task board (using the four \texttt{team\_task\_*} tools with revision checks on updates). When intervention is needed, only the lead can interrupt a teammate's current turn through \texttt{interrupt\_agent}.
Once the required work is complete, the lead reviews and tests the combined changes and produces the final response.

\textbf{Training.} We train Agent Team mode with an RL reward combining task performance, a collaboration bonus that encourages delegation and inter-agent communication, and a derived-latency penalty that promotes efficient coordination. Derived latency is computed by representing execution events and their collaboration dependencies as a directed acyclic graph (DAG), assigning costs from token counts at fixed prefill/decode rates plus measured tool-execution time, and taking the length of the critical path. This encourages useful parallelism while penalizing unnecessary sequential work and synchronization, with reduced sensitivity to serving-side batching and queuing delays.

\textbf{Performance.}
We construct a high-confidence subset of ProgramBench~\citep{yang2026programbenchlanguagemodelsrebuild} by retaining only tasks for which the reference solution achieves a pass rate of at least 95\% on the hidden test suite. This filtering procedure leaves 172 "golden" tasks. We also evaluate on FrontierSWE v2~\citep{proximal2026frontierswev2}, a larger and more challenging successor to FrontierSWE that uses a substantially improved methodology. We construct a no-GPU subset from the currently public tasks by excluding tasks that require GPU access. We evaluate single-agent and multi-agent configurations on both benchmarks under explicit per-rollout wall-clock deadlines. The results reported here are preliminary: we compare the strongest observed multi-agent configurations with the strongest available single-agent baselines.
On ProgramBench, we run up to three rollouts per task, corresponding to 516 planned rollouts for each configuration. We report Almost@1, which measures the fraction of individual rollouts achieving a score of at least 0.95. On FrontierSWE v2, we report Mean@5. ProgramBench deadlines range from 1 to 12 hours, while FrontierSWE v2 is evaluated at deadlines ranging from 1 to 20 hours. At each deadline, metrics are computed from the outputs available when the deadline is reached.

As shown in Figure \ref{fig:multi-agent-programbench}, multi-agent configurations outperform their single-agent counterparts at every deadline on both benchmarks. On ProgramBench, Almost@1 increases from 13.59\% at 1 hour to a peak of 30.04\% at 8 hours for the multi-agent configuration, compared with 12.79\% and 20.39\% for the single-agent configuration. On FrontierSWE v2, Mean@5 increases from 13.50\% at 1 hour to 32.90\% at 20 hours for the multi-agent configuration, compared with an increase from 10.50\% to 28.20\% for the single-agent configuration.

\section{Conclusion, Limitations, and Future Directions}
\label{sec:conclusion}

In this work, we introduce DeepSeek-V4.1-Flash, a multimodal Mixture-of-Experts (MoE) model with support for contexts of up to one million tokens. Through joint optimization of model architecture, cache precision, and deployment strategy, DeepSeek-V4.1-Flash pushes the limits of KV cache compression. Its Causal Encoder-Decoder (CED) architecture enables the model to activate only 8B parameters per token during prefill, compared with 16B during decode, improving cost efficiency for input-heavy agentic workloads. At equal sequence lengths, cross-layer KV cache reuse in Compressed Sparse Attention 2 (CSA2) and FP4 KV caching reduce its global KV cache footprint (always in HBM) to 890 bytes per token, roughly 1/4 of the corresponding footprint of DeepSeek-V4-Flash. SWA Bounded Replay further reduces its persistent KV cache footprint (always on SSD or in host memory) to roughly 1/8 of that of DeepSeek-V4-Flash. These reductions alleviate HBM and SSD capacity pressure while the model delivers substantially better overall performance than DeepSeek-V4-Flash. Despite possessing a significantly smaller parameter footprint than contemporary open-source models such as GLM-5.3 and Kimi-K3, DeepSeek-V4.1 achieves comparable—and in several tasks, superior—performance across key benchmarks.

Although DeepSeek-V4.1-Flash substantially simplifies several architectural components relative to DeepSeek-V4-Flash, the newly introduced architectural changes also create robustness boundaries that have yet to be fully characterized. Our internal evaluations cover a diverse range of test cases and boundary conditions, and we have not observed any systematic degradation in model capabilities in the evaluated settings. Nevertheless, no finite test suite can cover every extreme input and deployment condition. Potential selection errors in CSA2 and approximate state reconstruction in SWA Bounded Replay may still cause capability degradation in untested boundary cases. Going forward, we will continue to expand our stress-testing and evaluation stack, with particular attention to sparse retrieval over long contexts and SWA state reconstruction at cache-resumption boundaries. We will also monitor real-world workloads, systematically characterize potential failure modes and robustness boundaries, and further improve model robustness under extreme conditions. 

As AI models achieve remarkable performance capabilities, standard evaluation benchmarks have increasingly reached saturation. While DeepSeek-V4.1-Flash demonstrates performance that closely approaches top-tier models like Fable-5 and GPT-6 Astra—offering a highly comparable user experience in daily applications—a performance gap remains on the most challenging tasks. Although benchmark scores show a narrow margin, this parity does not imply that the model matches the frontier capabilities of leading closed-source systems on complex, high-difficulty reasoning and edge cases.

Consequently, we will continuously update our evaluation protocols to ensure rigorous assessment of state-of-the-art reasoning boundaries. Alongside continued efforts to reduce model costs, we believe that further advances in model intelligence will depend on the coordinated scaling of data, model capacity, and RL. With DeepSeek-V4.1-Flash as a new starting point, we will continue to explore the limits of model capabilities and systematically address key challenges in large-scale data synthesis and RL scaling. We will also actively integrate model–harness co-design, enabling the joint system to evolve and be optimized together. By advancing cost reduction and capability scaling in tandem, we hope to make highly capable agents more accessible and easier to deploy, further lowering the barriers to adopting AI technologies across a broader range of industries and scenarios.

\bibliography{main}

\newpage
\appendix

\section*{\LARGE Appendix}

\section{Author List}

Authors are listed alphabetically by their first name. 
Names marked with * denote individuals who have departed from our team. 

\noindent
\textbf{Research \& Engineering:} 
Anyi Xu,
B. Li,
Bangcai Lin,
Bing Xue,
BingCheng Xian,
Bingzheng Xu,
Bochao Wu,
Bowei Zhang,
Boyi Deng,
C.C. Yu,
Chao Jin,
Chaofan Lin,
Chen Dong,
Chenbing Wang,
Chenfan Feng,
Chengda Lu*,
Chenggang Zhao,
Chengqi Deng,
Chengyuan Zhang,
Chenhao Xu,
Chenqi Zhao,
Chenze Shao,
Chuhao Wang,
Chuqi Zhang,
Damai Dai,
Dejian Yang,
Deli Chen,
Di Huang,
Di Wu,
Donghao Li,
Erhang Li,
Eric Fu,
F. Zhou,
Fangwei Zhou,
Fangyun Lin,
Fangzhou Yuan,
Feiyu Xia,
Fucong Dai,
Guangbo Hao,
Guanglin Li,
Guanting Chen*,
Guoai Cao,
Guofan Fan,
Guolai Meng,
Guowei Li,
Haichuan Zhang,
Haiyang Ma,
Haiyang Shen,
Han Li,
Han Yu,
Han Zhang,
Hangyuan Deng,
Hanwei Xu,
Hanxiang Xu,
Hanxun Zhong,
Hao Guo,
Hao Jiang,
Hao Li,
Hao Qin,
Haodong Wen,
Haofen Liang,
Haofeng Huang,
Haohua Liu,
Haoling Zhang,
Haoming Luo,
Haoran Yang,
Haotian Xu*,
Haotian Yuan,
Haoting Huang,
Haowen Luo,
Haoyang Cai,
Haoyu Chen,
Haozhe Ji,
Hengran Zhang,
Hengrui Wang,
Hengxu Wu,
Honghui Ding,
Hongxuan Tang,
Huadong Wang,
Huanqi Cao,
Huazuo Gao,
Hui Qu,
Hui Zeng,
J. Yang,
J.H. Jin,
J.H. Zhang,
J.X. Zou,
Jia Yu,
Jiahui Zhou,
Jiajun Chen,
Jialiang Huang,
Jialin Zhao,
Jiamin Tang,
Jian Zhou,
Jianan Tong,
Jianwen Li,
Jiaqi Zhu,
Jiarui Wang,
Jiasheng Ye,
Jiashi Li,
Jiaxin Xu,
Jiaying Ding,
Jibai Lu,
Jiewen Hu,
Jin Yan,
Jincheng Zhai,
Jingchang Chen,
Jingcheng Hu,
Jingli Zhou,
Jingsheng Xu,
Jingting Xiang,
Jingyan Yun,
Jingyang Yuan,
Jingyuan Cheng,
Jinhua Zhu,
Jinpeng Wang,
Jinyi Chen,
Jinyi Hu,
Jiping Yu,
Jueliang Guo,
Junbo Pei,
Junbo Sun,
Junguang Jiang,
Junjie Qiu,
Junkang Zhou,
Junqi Liu,
Junren Li,
Junxian Li,
Junxiao Song,
Junyi Guo,
Kai Dong,
Kaifeng Chen,
Kaige Gao,
Kang Guan,
Kangdong Yuan,
Ke Hong,
Ke Xu,
Kefan Zhao,
Kexin Ji,
Kexin Zhang,
Kexing Zhou,
Kuai Yu,
Lan Zhang,
Lean Wang,
Lecong Zhang,
Lei Wang,
Letian Gao,
Liang Zhao,
Liansheng Xu,
Lihua Guo,
Lingxiao Luo,
Lingyue Fu,
Litao Deng,
Litong Wang,
Liyue Zhang,
Longhao Chen,
Lu Chen,
Luotian Huang,
Luyao Ma,
Luyao Wang,
M.S. Di,
Max Mei,
Menghao Ye,
Miao Cui,
Mingchuan Zhang,
Minghua Zhang*,
Minghui Tang,
Mingjing Zhang,
Mingqi Wei,
Mingshu Chen,
Mingxing Liu,
Mingxu Zhou,
Mingyu Xu,
Mingyu Yang,
Mingze Wang,
Muyang Chen,
Ni Shentu,
Ning Wang,
Niufang Ning,
Panpan Huang,
Peixin Cong,
Peiyi Wang,
Peiyuan Xin,
Pengfei Ren,
Pengfei Yan,
Pengle Zhang,
Qi Kang,
Qi Tang,
Qiancheng Wang,
Qiang Li,
Qihao Zhu,
Qingyang Li,
Qinyu Chen,
Qiushi Du,
Qizhou Guo,
Rongxian Xu,
Rui Ding,
Rui Hu,
Rui Tian,
Rui Yu,
Ruidong Zhu,
Ruifan Xu,
Ruihan Yang,
Ruihang Xia,
Ruijie Lu,
Ruilin Geng,
Ruipeng Hong,
Ruiqi Ge,
Ruisong Zhang,
Ruize Sun,
Ruizhe Pan,
Runji Wang,
Runqian Chen,
Runxin Xu,
Ruohong Tian,
Ruomeng Shen,
Ruoyu Zhang,
Ryan X.,
S.H. Liu,
Shanghao Lu,
Shangyan Zhou,
Shanhuang Chen,
Shaofei Cai,
Shaoheng Nie,
Shaoyuan Chen,
Shengding Hu,
Shengkai Lin,
Shengwen Ran,
Shengyu Liu,
Shengyuan Jia,
Shi Bai,
Shi Feng,
Shicheng Xu,
Shichun Liu,
Shiqiang Hu*,
Shirong Ma,
Shiyu Wang,
Shiyuan Feng,
Shufan Gong,
Shuhan Lin,
Shuiping Yu,
Shunfeng Zhou,
Shuo Yang,
Shuomeng Wang,
Shuting Guo,
Shuting Pan,
Shuying Yu,
Sinuo Cao,
Siyi Lin,
Sizhe Chen,
Songyang Chen,
Songyang Zhou,
Tao Ni,
Tao Yun,
Tian Jin,
Tian Pei,
Tian Ye,
Tianle Lin,
Tianran Ji*,
Tianyi Cui,
Tianyuan Yue,
Tingting Yu,
Tongrui Xiong,
Wangding Zeng,
Wei Liu,
Wei Zhang,
Weibin Xu,
Weihao Zeng,
Weilin Zhao,
Wen Liu,
Wenfeng Liang,
Wenjie Pang,
Wenjing Luo,
Wenjing Yao*,
Wenjun Gao,
Wenkai Shao,
Wenkai Yang,
Wenli Zhang,
Wenlu Wang,
Wenlve Huang,
Wenqian Yan,
Wentao Zhang,
Xi Gao,
Xiang He,
Xiang Li,
Xiangli Li,
Xiangwen Wang,
Xiangying Zhang,
Xiankui Wei,
Xiao Bi,
Xiaodong Liu,
Xiaohan Wang,
Xiaojian Qu,
Xiaokang Chen,
Xiaokang Zhang,
Xiaotao Nie,
Xiaoyao Zou,
Xiaoyuan Li,
Xicheng Guo,
Xieting Chu,
Xin Cheng,
Xin Liu,
Xin Xie,
Xinbo Xu,
Xingchao Liu,
Xingchen Liu,
Xingkai Yu,
Xingyou Li,
Xintong Yao,
Xinyang Chen,
Xinyong Jiang,
Xinyu Yang,
Xinyu Yang,
Xu Chen,
Xuanyu Wang,
Xubei Zhong,
Xuecheng Su,
Xuejie Liu,
Xuheng Lin,
Xujie Fan,
Xuncheng Zhao,
Xuwei Fu,
Y.C. Yan,
Y.H. Jiang,
Y.T. Wu*,
Y.W. M.,
Y.Z. Wang,
Yafei Gao,
Yang Yang,
Yang Zhang,
Yanru Ma,
Yanwen Huang,
Yao Li,
Yao Li,
Yao Meng,
Yao Zhao,
Yaofeng Sun,
Yaohui Wang,
Yaoyang Ye,
Yehang Yin,
Yexinrui Wu,
Yi Qian,
Yi Tao,
Yi Yu,
Yichao Zhang,
Yichen Jiang,
Yicheng Wang,
Yifan Ding,
Yifan Shi,
Yifeng Peng,
Yifeng Zhai,
Yijia Wu,
Yiliang Xiong,
Yilun Wang,
Ying He,
Ying Zhou*,
Yingjia Luo,
Yinmin Zhong,
Yiping Wang,
Yisong Wang,
Yixiang Zhang,
Yixiao Chen,
Yixuan Tan,
Yixuan Wei,
Yiyang Ma,
Yiyao Yang,
Yiyuan Liu,
Yizai Cai,
Yizhen Wei,
Yizhi Wang,
Yonglun Yang,
Yongqi Zhuo,
Yongqiang Guo,
Yongtong Wu,
Yu Wu,
Yu Zhang,
Yuan Bian,
Yuan Cheng,
Yuan Ou,
Yuan Sun,
Yuanfan Xu,
Yuanhang Sun,
Yuanhao Li,
Yuchen Liu,
Yuchen Yao,
Yudong Han,
Yuduan Wang,
Yuhan Wu,
Yuhao Meng,
Yuheng Zou,
YuKun Li,
Yunchuan Wang,
Yunfan Xiao,
Yunfan Xiong,
Yupeng Chen,
Yuqian Cao,
Yuqian Wang,
Yuqing Chen,
Yushun Zhang,
Yutong Lin,
Yuwei Xiao,
Yuxian Gu,
Yuxiang Chen,
Yuxiang Huang,
Yuxiang Luo,
Yuxiang You,
Yuxin Chen,
Yuxin Xiang,
Yuxuan Liu,
Yuxuan Zhou,
Yuyang Zhou,
Yuzhe Guo,
Yuzhen Huang,
Yuzhuo Bai,
Z.Y. Z.,
Zanlin Ni,
Zehao Wang,
Zehua Zhao,
Zehui Ren,
Zejun Zhao,
Zhangli Sha,
Zhanying Wang,
Zhaochen Zhang,
Zhaoshuai Du,
Zhe Fu,
Zhean Xu,
Zhenda Xie,
Zheng Liu,
Zhengyan Zhang,
Zhenhua Dong,
Zhewen Hao,
Zhibang Wang,
Zhibin Gou,
Zhicheng Ma,
Zhihao Li,
Zhihong Shao,
Zhihuan Huang,
Zhijie Li,
Zhirui Lu,
Zhixian Huang,
Zhixuan Chen,
Zhixuan Chen,
Zhixuan Pan,
Zhiyu Wu,
Zhizhou Ren,
Zhu He,
Zhuoshu Li,
Zhuping Zhang,
Zian Xu,
Zihao Wang,
Zihui Gu,
Zijia Zhu,
Zili Zhang,
Zilin Li,
Zilong Hou,
Zilong Lyu,
Ziqiao Wang,
Ziwei Xie,
Ziya Zhang,
Ziyi Gao,
Zizheng Pan,
Zonglin Li,
Zongqing Yao,
Zui Chen,
Zuofan Wu

\noindent
\textbf{Business \& Compliance:}
Chenchen Ling,
Chengyu Hou,
Chong Chen,
D. Li,
Di Qi,
Dongjie Ji,
Fang Wei,
Fanyi Xia,
Fei Xie,
Feiyi Tan,
Hailong Guo,
Haiyan Zhai,
Hui Zhou,
Huihui Tan,
Huijie Li,
Jia Luo,
Jia Song,
Jialu Cai,
Jian Liang,
Jiangting Zhou,
Jiaqi Gao,
Jiayi Shao,
Jie Chen,
Jieyu Yang,
Jin Chen,
Jingde Zhang,
Jingzi Zhou,
Jinqian Wang,
Jinyang Liu,
JinZhao Sun,
Junhua Ling,
Junmin Zheng,
Kaicheng Yang,
Ke Xu,
Le Su,
Leyi Xia,
Liangfeng Ding,
Lin Zhuo,
Linwang Ma,
Linyan Zhu,
Liyu Cai,
Luqi Yao,
M.K. Zhang,
Meng Li,
Miao Lin,
Miaojun Wang,
Min Zhang,
Mingming Li,
Mingming Wang,
Mingze Yin,
Minmin Han,
Nan Cao,
Ning Wang,
Ningxin Ma,
Panpan Wang,
Peihan Lin,
Peng Sun,
Peng Zhang,
Qian Ying,
Qiang Xiang,
Qiao Wang,
Qingmiao Mao,
Qiwei Jiang,
Rongli Jin,
Ruyi Chen,
Sha Tao,
Shangmian Sun,
Shaoqing Wu,
Shichao Zou,
Si Lei,
Tianyang Zhang,
Tianyu Sun,
Tingting Yin,
W.L. Xiao,
Wei An,
Wei Li,
Wei Wang,
Weiwei Lin,
Wenqing Hou,
X. Lin,
Xiangfei Meng,
Xianzhu Huang,
Xiao Peng,
Xiaoqian Li,
Xiaoting Zhang,
Xiaowen Sun,
Xiaoxiang Wang,
Xiaoyu Ye,
Xinrou Zhang,
Xinyu Zhang,
Xue Cao,
Xueyin Chen,
Yanan Zhou,
Yanhong Xu,
Yao Xia,
Yao Xu,
Yi Shao,
Yihong Zhang,
Yiling Ma,
Ying Tang,
Yining Lou,
Yiru Chen,
Yishi Piao,
Yixuan Chen,
Yong Xiong,
Yuchen Xuan,
Yuehan Yang,
Yuer Xu,
Yukun Zha,
Yunxian Ma,
Yuping Lin,
Yuting Yan,
Yutong Xie,
Yuwen Sheng,
Yuxuan Zhu,
Zekai Zhang,
Zhe Ju,
Zhenzhen Lin,
Zheren Gao,
Zheyang Sun,
Zhigang Yan,
Zhongyu Wu,
Zi Wang,
Zihua Qu,
Ziling Yan,
Ziyi Wan

\section{Evaluation Details}

\subsection{Scaffold Configurations}
\label{app:multi-scaffold-eval}

\begin{table}[ht]
  \centering
  \footnotesize
  \setlength{\tabcolsep}{3pt}
  \renewcommand{\arraystretch}{1.0}
  \caption{\centering Performance across Claude Code versions at Max reasoning effort.}
  \label{tab:claude-code-versions}
  \newsavebox{\claudecodeversionstable}
  \begin{lrbox}{\claudecodeversionstable}
  \begin{tabular}{@{}l|cccc|c@{}}
    \toprule
    \textbf{Benchmark {\tiny (Metric)}}
      & \textbf{v2.1.105} & \textbf{v2.1.238} & \textbf{v2.1.251} & \textbf{v2.1.259}
      & \textbf{Average} \\
    \midrule
    DeepSWE v1.1 {\tiny (Resolved)}
      & 68.4 & 68.7 & 69.8 & 68.6 & 68.9 \\
    Terminal-Bench v2.1 {\tiny (Pass@1)}
      & 87.3 & 88.4 & 88.0 & 87.6 & 87.8 \\
    \bottomrule
  \end{tabular}
  \end{lrbox}
  \usebox{\claudecodeversionstable}
  \par\vspace{3pt}
  \begin{minipage}{\wd\claudecodeversionstable}
    \scriptsize
    \textit{Note.} Average is computed from the unrounded Pass@1 scores of the four versions.
  \end{minipage}
\end{table}

All scaffolds run in Linux task containers using the shared evaluation settings in Table~\ref{tab:agent-scaffolds}.
Each run starts from the benchmark task description and uses the prompts, task templates, and tool definitions supplied by the scaffold's runtime or evaluation integration.
We add no experimental system prompt.

\begin{itemize}[leftmargin=1.5em, itemsep=3pt, topsep=3pt, parsep=0pt]
  \item \textbf{Claude Code (v2.1.105, v2.1.238, v2.1.251, v2.1.259).}
  We use the Claude Agent SDK with each version's native tool interface.
  Table~\ref{tab:agent-scaffolds} reports v2.1.251; Table~\ref{tab:claude-code-versions} compares all four versions.

  \item \textbf{Codex (v0.147.0).}
  We use standard app-server mode with adapted tool schemas.

  \item \textbf{OpenCode (v1.18.15).}
  We use the \texttt{build} agent with shell and file tools, and native task delegation.

  \item \textbf{Pi (v0.84.2).}
  We use RPC mode with file and shell tools plus a search extension exposing \texttt{search}, \texttt{open\_page}, and \texttt{find\_in\_page}.

  \item \textbf{mini-SWE.}
  We use the \texttt{mini\_swe\_v2} port\footnote{\href{https://github.com/SWE-agent/mini-swe-agent/tree/04d809ceab9df28f9adaed044884180159172930}{mini-swe-agent, commit \texttt{04d809ceab9d}}.} with a single \texttt{bash} tool, requiring a tool call each turn and a submission marker to finish.

  \item \textbf{DeepSeek Harness (DSH).}
  We use \emph{Minimal} with a single \texttt{bash} tool;
  \emph{Standard} with the full \texttt{sdk} profile and 26 initial function tools, including web search and fetch;
  and \emph{PTC} with \texttt{run\_code} for TypeScript programs using 24 underlying tools.
  Standard and PTC use v0.1.1+\allowbreak custom.\allowbreak 202609011522.
\end{itemize}

\subsection{Reasoning Efforts across Scaffolds}
\begin{figure}[t]
    \centering
    \includegraphics[width=0.9\linewidth]{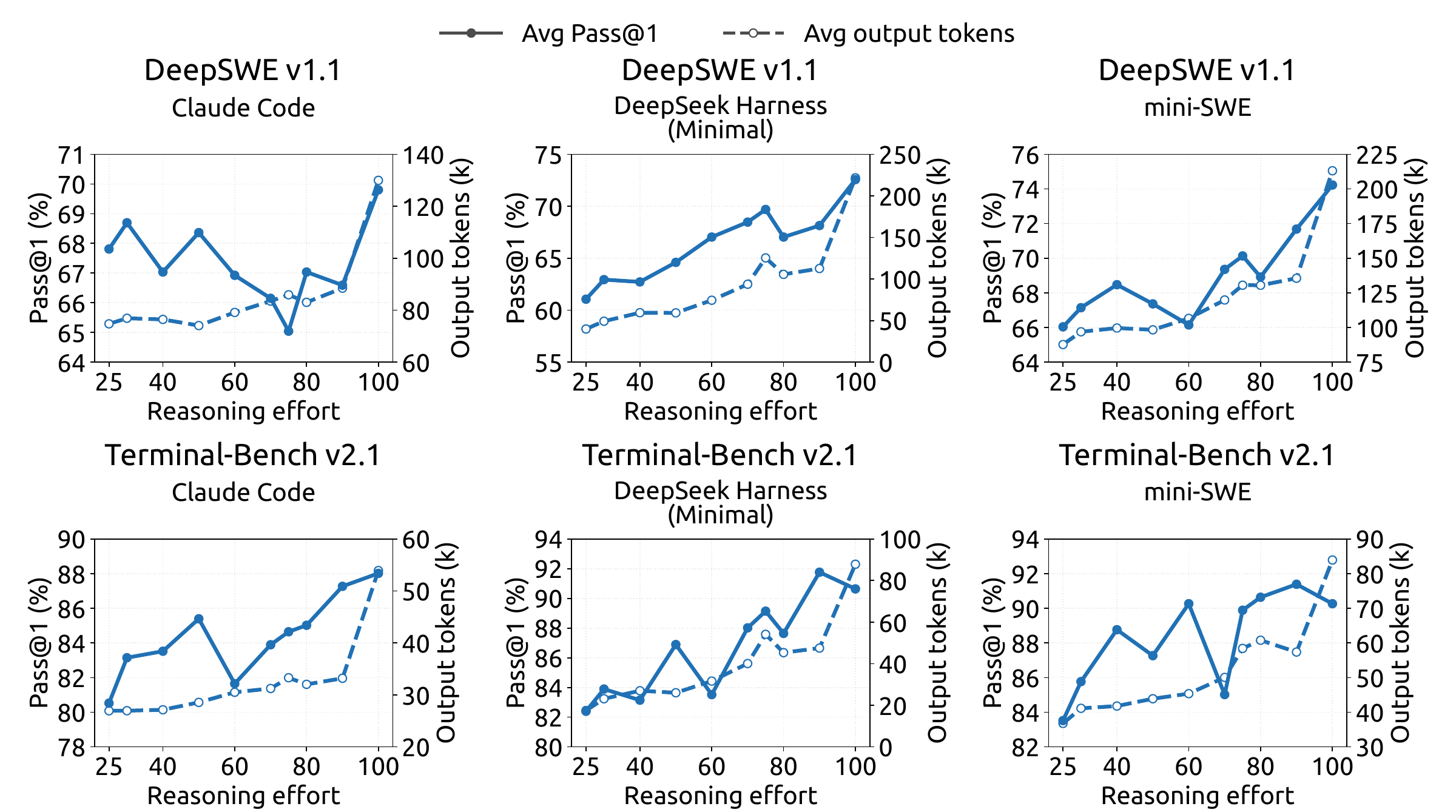}
    \caption{Reasoning effort drives trajectory length consistently but
    correlates only weakly with accuracy across coding scaffolds.
    Each panel plots Pass@1 (\%, solid, left axis) and mean output tokens
    per trajectory (k, dashed, right axis) against the reasoning-effort
    setting, for DeepSWE v1.1 (top row) and Terminal-Bench v2.1 (bottom row) under three agent scaffolds: Claude Code,
    DeepSeek Harness (Minimal) and mini-SWE. All panels come from the same
    checkpoint. 
}
    \label{fig:reasoning-efforts-scafolds}
\end{figure}

Across all six panels in Figure~\ref{fig:reasoning-efforts-scafolds}, raising the reasoning-effort setting lengthens the
trajectories: mean output tokens per trajectory grow monotonically with
effort in every scaffold--benchmark pair. Pass@1 tracks this growth only
loosely. It improves overall, but the response is not monotone, with
plateaus and dips at intermediate settings in most panels. The three
scaffolds are also calibrated differently: on DeepSWE v1.1, Claude Code
is the flattest curve and spends comparatively few extra tokens, whereas
DeepSeek Harness (Minimal) starts lowest and gains the most at the largest
token cost, with mini-SWE in between. On Terminal-Bench v2.1 the three
scaffolds are compressed into a narrow band, and the ordering at maximum
effort favours DeepSeek Harness (Minimal), followed by mini-SWE and Claude
Code: scaffold choice matters at least as much as the effort tier
once the task is nearly saturated.

\subsection{Detailed Results of Reasoning Benchmarks across Reasoning Efforts}
\begin{figure}[t]
    \centering
    \includegraphics[width=1.0\linewidth]{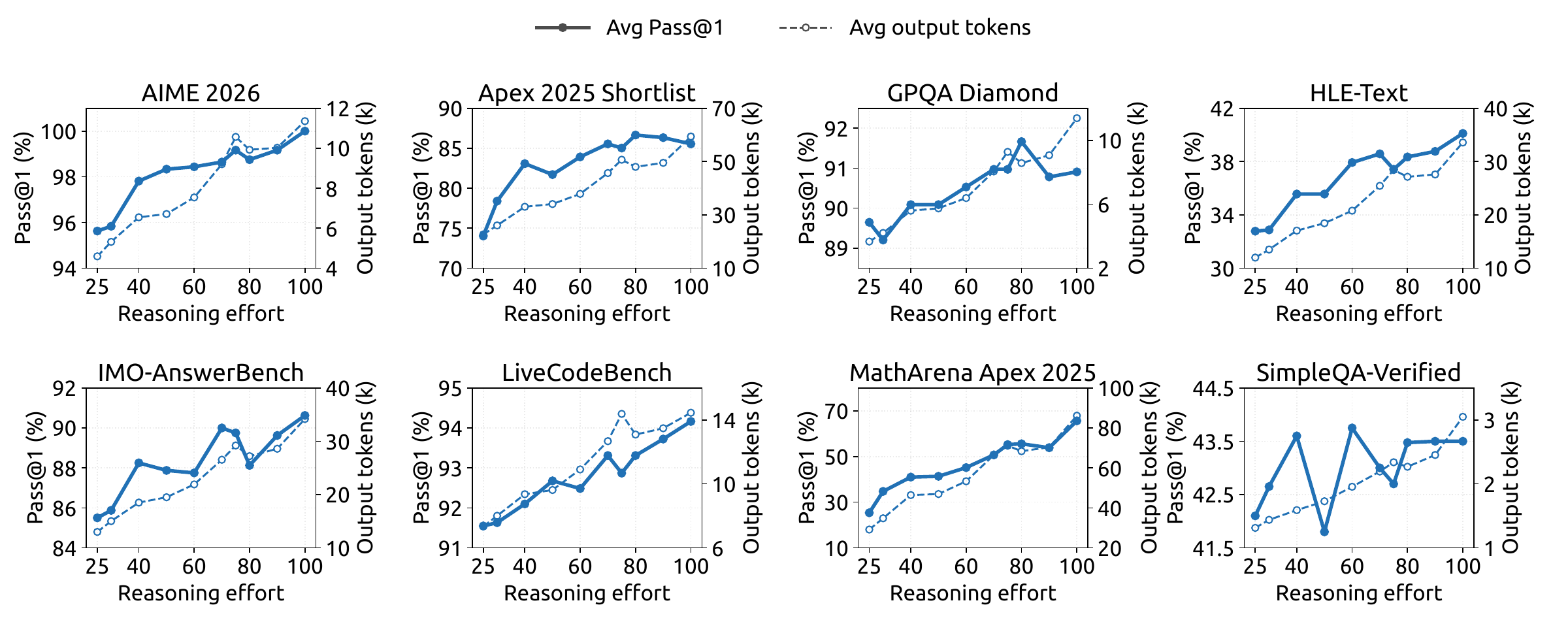}
    \caption{Performance and output length as a function of reasoning effort on eight reasoning-intensive benchmarks. Each panel plots
Pass@1 (solid, left axis) and mean output tokens per response (dashed, right axis) as the
reasoning-effort value is varied from 25 to 100.}
    \label{fig:effort_reasoning_benchmarks}
\end{figure}

Figure~\ref{fig:effort_reasoning_benchmarks} shows that the reasoning-effort setting gives
smooth, well-behaved control over both output length and accuracy, consistently across all
eight benchmarks, which span competition mathematics, science QA, open-domain knowledge and
code. On the length side, raising the effort from 25 to 100 scales the average response
predictably on every benchmark -- a uniform $2.0$--$3.1\times$ increase, from 4.6k to 11.4k
tokens per response on AIME 2026 and from 29.1k to 86.1k on MathArena Apex 2025 -- with no
runaway growth or anomalies, so the compute cost of any tier can be estimated in advance.
Accuracy follows the same smooth trajectory and responds in the right direction on every
benchmark, with no benchmark ever degrading as the effort increases: MathArena Apex 2025
gains $+40.3$ points (25.3\%$\to$65.6\%) and Apex 2025 Shortlist $+11.5$, while even the
already-saturated benchmarks remain stable (GPQA Diamond $+1.3$, LiveCodeBench $+2.6$), and
AIME 2026 reaches a full 100\%. The model thus exposes a single, reliable knob that moves
the cost--accuracy operating point in a controlled and predictable way, allowing each
deployment to select the effort tier that matches its latency and compute budget without
sacrificing accuracy.

\section{Exponential Token Penalty in Reasoning Effort Control}
\label{app:effort-penalty}

The scalar effort variable provides a deployment-time control over the
cost--quality trade-off without imposing a hard token budget. During
reinforcement learning, lower effort levels apply a stronger token penalty,
whereas higher effort levels permit more computation. This section gives a
simplified motivation for the exponential penalty schedule.

For a trajectory with $\ell$ reasoning tokens generated at effort level $b$,
the length deduction is
\begin{equation}
    r^{\mathrm{len}}(\ell,b)
    =
    -\min\left\{
        C_{\max},
        k(b)\frac{\ell}{L_{\mathrm{norm}}}
    \right\},
    \label{eq:app-length-penalty}
\end{equation}
where $L_{\mathrm{norm}}$ is a reference length and $C_{\max}$ caps the
deduction. The effort-dependent token-penalty coefficient is
\begin{equation}
    k(b)
    =
    k_0\exp\left(-\frac{b-b_{\min}}{\tau}\right),
    \label{eq:app-exp-penalty}
\end{equation}
where $k_0$ is the penalty coefficient at the lowest effort level $b_{\min}$
and $\tau$ controls the rate of penalty decay.

To motivate this choice, consider a fixed problem $x$. Let $p_x(\ell)$ denote
its probability of being solved after $\ell$ reasoning tokens. In the uncapped
region, define the preferred reasoning length $\ell_x^*(b)$ by
\begin{equation}
    \ell_x^*(b)
    \in
    \arg\max_{\ell\geq 0}
    \left[
        p_x(\ell)
        -
        k(b)\frac{\ell}{L_{\mathrm{norm}}}
    \right].
    \label{eq:app-optimal-length}
\end{equation}
For an interior optimum, the first-order condition is
\begin{equation}
    p_x'\!\left(\ell_x^*(b)\right)
    =
    \frac{k(b)}{L_{\mathrm{norm}}},
    \label{eq:app-marginal-balance}
\end{equation}
where $p_x'(\ell)=d p_x(\ell)/d\ell$ is the marginal improvement in solve
probability from additional reasoning.

We assume that this marginal benefit decays approximately exponentially over
the relevant operating range:
\begin{equation}
    p_x'(\ell)
    \approx
    a_x\exp\left(-\frac{\ell}{s_x}\right),
    \label{eq:app-marginal-decay}
\end{equation}
where $a_x>0$ is an instance-dependent scale and $s_x>0$ determines the decay
rate. Substituting Eqs.~\eqref{eq:app-exp-penalty} and
\eqref{eq:app-marginal-decay} into Eq.~\eqref{eq:app-marginal-balance} gives
\begin{equation}
    \ell_x^*(b)
    \approx
    C_x-s_x\log k_0
    +
    \frac{s_x}{\tau}(b-b_{\min}),
    \label{eq:app-effort-length-trend}
\end{equation}
where $C_x=s_x\log(a_xL_{\mathrm{norm}})$ is independent of $b$. Thus, the
exponential penalty schedule produces a simple first-order affine trend
between the requested effort and the preferred reasoning length under this
local model.

For two effort levels $b_2>b_1$, the corresponding predicted length difference
is
\begin{equation}
    \ell_x^*(b_2)-\ell_x^*(b_1)
    \approx
    \frac{s_x}{\tau}(b_2-b_1).
    \label{eq:app-effort-separation}
\end{equation}
Consequently, $k_0$ mainly controls the overall pressure toward shorter
reasoning, while $\tau$ controls the predicted sensitivity to effort.

This derivation is a local reward-level approximation, not a claim that
measured average lengths must be linear or pointwise monotonic. Realized
behavior may deviate because the effort instruction can directly change the
reasoning strategy, generation is stochastic, agent trajectories contain
different numbers of turns, and subgroup reward normalization changes
optimization strength. The analysis also assumes an interior solution for
which the penalty cap is inactive. Once the cap is reached, the marginal token
penalty becomes zero and the capped region must be considered separately.

\end{CJK*}
\end{document}